\documentclass[draftclsnofoot,twocolumn]{IEEEtran}

\usepackage{graphicx}
\usepackage{bm}
\usepackage{subfigure}
\usepackage{epsfig}
\usepackage{latexsym}
\usepackage{amsmath}
\usepackage{amsthm}
\usepackage{amssymb}
\usepackage{amsfonts}
\usepackage{color}
\usepackage{enumerate}
\usepackage{multirow}
\usepackage{booktabs}
\usepackage{epstopdf}
\usepackage{algorithm}
\usepackage{algpseudocode}
\usepackage{threeparttable}
\usepackage{diagbox}
\usepackage{lettrine}
\usepackage{colortbl}
\usepackage{booktabs}
\usepackage{graphicx}
\usepackage{setspace}
\usepackage{algorithmicx,algorithm}
\usepackage{url}
\usepackage[utf8]{inputenc}

\definecolor{mycH1}{rgb}{0,0,0.8125}
\definecolor{mycH2}{rgb}{0,0.0625,1}
\definecolor{mycH3}{rgb}{0,0.3125,1}
\definecolor{mycH4}{rgb}{0,0.6250,1}
\definecolor{mycH5}{rgb}{0,0.8750,1}
\definecolor{mycH6}{rgb}{0.1250,1,0.8750}
\definecolor{mycH7}{rgb}{0.3750,1,0.6250}
\definecolor{mycH8}{rgb}{0.6875,1,0.3125}
\definecolor{mycH9}{rgb}{0.9375,1,0.0625}
\definecolor{mycH10}{rgb}{1,0.8125,0}
\definecolor{mycH11}{rgb}{1,0.5625,0}
\definecolor{mycH12}{rgb}{1,0.2500,0}
\definecolor{mycH13}{rgb}{1,0,0}
\definecolor{mycH14}{rgb}{0.7500,0,0}
\definecolor{mycH15}{rgb}{0.5000,0,0}
\definecolor{mycT1}{rgb}{0,0.2000,1}
\definecolor{mycT2}{rgb}{0,1,1}
\definecolor{mycT3}{rgb}{0.6000,1,0.4000}
\definecolor{mycT4}{rgb}{1,0.800,0}
\definecolor{mycT5}{rgb}{1,0,0}
\definecolor{mycT6}{rgb}{0.6000,0,0}

\ifCLASSINFOpdf
\else
\fi
\begin{document}
\begin{spacing}{1.0}

\begin{center}
\LARGE
\textbf{$A^{3}$CLNN: Spatial, Spectral and Multiscale Attention ConvLSTM Neural Network for Multisource Remote Sensing Data Classification}

\end{center}

~\\

\textcircled{c} 2022 IEEE.  Personal use of this material is permitted.  Permission from IEEE must be obtained for all other uses, in any current or future media, including reprinting/republishing this material for advertising or promotional purposes, creating new collective works, for resale or redistribution to servers or lists, or reuse of any copyrighted component of this work in other works.

~\\

\textbf{DOI:}\textcolor[rgb]{0.00,0.00,1.00}{\underline{10.1109/TNNLS.2020.3028945}}

~\\

Heng-Chao~Li, \emph{Senior Member, IEEE}, Wen-Shuai~Hu, Wei~Li, \emph{Senior Member, IEEE}, Jun~Li, \emph{Senior Member, IEEE}, Qian~Du, \emph{Fellow, IEEE}, and Antonio Plaza, \emph{Fellow, IEEE}

\newpage
\title{$A^{3}$CLNN: Spatial, Spectral and Multiscale Attention ConvLSTM Neural Network for Multisource Remote Sensing Data Classification}
%
%
%

\author{
\thanks{Manuscript received February 7, 2020; revised August 25, 2020; accepted September 24, 2020. Date of publication October 21, 2020; date of current version February 4, 2022. This work was supported by the National Natural Science Foundation of China under Grant 61871335, in part by the Fundamental Research Funds for the Central Universities under Grant 2682020XG02 and 2682020ZT35. (Corresponding author: Wen-Shuai~Hu.)}
        Heng-Chao~Li, \emph{Senior Member, IEEE}, Wen-Shuai~Hu,
        Wei~Li, \emph{Senior Member, IEEE}, Jun~Li, \emph{Senior Member, IEEE}, Qian~Du, \emph{Fellow, IEEE}, and Antonio Plaza, \emph{Fellow, IEEE}
\thanks{Heng-Chao Li and Wen-Shuai Hu are with the Sichuan Provincial Key Laboratory of Information Coding and Transmission, Southwest Jiaotong University, Chengdu 611756 China (e-mail: lihengchao\_78@163.com; wshu@my.swjtu.edu.cn).}
\thanks{Wei Li is with the School of Information and Electronics, Beijing Institute of Technology, Beijing 100081 China.}
\thanks{Jun Li is with the Guangdong Provincial Key Laboratory of Urbanization and Geo-simulation, School of Geography and Planning, Sun Yat-sen University, Guangzhou 510275 China.}
\thanks{Qian Du is with the Department of Electrical and Computer Engineering, Mississippi State University, Mississippi State, MS 39762 USA.}
\thanks{Antonio Plaza is with the Hyperspectral Computing Laboratory, Department of Technology of Computers and Communications, Escuela Polit$\acute{e}$cnica, University of Extremadura, 10060 C$\acute{a}$ceres, Spain.}

}

%
%

\markboth{IEEE TRANSACTIONS ON NEURAL NETWORKS AND LEARNING SYSTEMS}%
{Shell \MakeLowercase{\textit{et al.}}: Bare Demo of IEEEtran.cls for Journals}
%



\maketitle

\begin{abstract}
The problem of effectively exploiting the information multiple data sources has become a relevant but challenging research topic in remote sensing. In this paper, we propose a new approach to exploit the complementarity of two data sources: hyperspectral images (HSIs) and light detection and ranging (LiDAR) data. Specifically, we develop a new dual-channel spatial, spectral and multiscale attention convolutional long short-term memory neural network (called dual-channel $A^{3}$CLNN) for feature extraction and classification of multisource remote sensing data. Spatial, spectral and multiscale attention mechanisms are first designed for HSI and LiDAR data in order to learn spectral- and spatial-enhanced feature representations, and to represent multiscale information for different classes. In the designed fusion network, a novel composite attention learning mechanism (combined with a three-level fusion strategy) is used to fully integrate the features in these two data sources. Finally, inspired by the idea of transfer learning, a novel stepwise training strategy is designed to yield a final classification result. Our experimental results, conducted on several multisource remote sensing data sets, demonstrate that the newly proposed dual-channel $A^{3}$CLNN exhibits better feature representation ability (leading to more competitive classification performance) than other state-of-the-art methods.
\end{abstract}

\begin{IEEEkeywords}
Multisource remote sensing data classification, convolutional long short-term memory, attention mechanism, transfer learning, feature extraction, fusion.
\end{IEEEkeywords}

%
\IEEEpeerreviewmaketitle

\section{Introduction}
%
%
%
%
\lettrine[lines=2]{W}{ith} the development of remote sensing technology, different sources of complementary data are now available from a variety of sensors. Hyperspectral images (HSIs) provide plenty of spectral information and have been widely used for land-cover classification purposes \cite{JM2013}-\cite{GC2014}. Different from HSI data, light detection and ranging (LiDAR) data consist of detailed elevation information. These data convey rich information in the spatial domain that can be used to improve the characterization of HSI scenes \cite{MB2014}-\cite{IT2015}, as the LiDAR data are less affected by atmospheric interferers \cite{JJ2014}. In the literature, several works \cite{WL2014}-\cite{BR2017} have discussed the fusion of multisource remote sensing data.

Many classification methods have been proposed to exploit the spatial-spectral information contained in HSI data, including machine learning-based methods \cite{LP2017}, \cite{WL2015}, tensor-based algorithms \cite{YD2018}, sparse representation-based methods \cite{MC2015}. In recent years, deep learning-based algorithms have achieved great success in remote sensing data interpretation. Convolutional neural networks (CNNs) were first adopted for HSI classification by Hu \emph{et al.} \cite{WH2015} and Chen \emph{et al.} \cite{YC2016}. After these seminal works, many other deep learning-based HSI classification methods have been proposed, and these methods have been shown to be able to provide higher classification accuracies than traditional methods. Relevant examples are the CNN-based pixel-pair model \cite{WL2017}, a spatial-spectral feature based classification (SSFC) model that stacks CNNs with balanced local discriminant embedding \cite{WS2016}, or the siamese CNN-based method \cite{BL2018}. In addition to the CNN-based HSI classification algorithms, recurrent neural networks (RNNs) \cite{AG2009} have also achieved great success in the task of capturing useful information from different kinds of inputs, due to their unique ability for modeling long-range dependencies. As a result, many HSI classification models have been developed, including the RNN-based pixel-level spectral classification model \cite{LM2017}, a local spatial sequential RNN (LSS-RNN) model \cite{XZ2018}, and a classification model that combines CNNs and RNNs \cite{RH2019}. Furthermore, in order to solve the gradient vanishing or explosion problem in RNNs, long short-term memories (LSTMs) \cite{SH1997} and convolutional LSTMs (ConvLSTMs) \cite{XS2015} were proposed. The most common way to utilize them is in combination with a CNN, such as the convolutional RNN (CRNN) model for spectral-contextual feature extraction \cite{HW2017}, a recurrent three-dimensional (3-D) fully convolutional network \cite{AS2018}, and a two-stage classification model that combines a 3-D CNN and and a (2-D) ConvLSTM \cite{MS2019}. In addition, there are also some relevant works focused on building deep feature extraction and classification models using the LSTM and (2-D) ConvLSTM cells as basic units, such as a spatial-spectral LSTMs (SSLSTMs) \cite{FZ2017}, the bidirectional-ConvLSTM (Bi-CLSTM) \cite{QL2017}, or a spatial-spectral ConvLSTM 2-D neural network (SSCL2DNN) \cite{WH2019}. These methods have achieved good performance in the task of HSI data classification. In order to better preserve the intrinsic structure of HSI data, a 3-D ConvLSTM cell was developed from the (2-D) ConvLSTM cell in \cite{WH2019}, from which a spatial-spectral ConvLSTM 3-D neural network (SSCL3DNN) was designed.

Considering the special characteristics of HSI and LiDAR data, several works aimed at fusing these two data sources in order to improve the classification performance \cite{BR2017}. Examples include decision-fusion classification methods \cite{WL2014}, \cite{CZ2016} and morphological feature extraction-based algorithms \cite{WL2013}-\cite{PG2016}.
In addition, several deep learning-based classification methods have also been proposed for the classification of multisource remote sensing data. Xu \emph{et al.} \cite{XX2018} built a two-branch CNN model with data augmentation for fusing HSI and LiDAR features. In \cite{MZ2020}, an unsupervised patch-to-patch CNN (PToP CNN) model was designed for HSI and LiDAR data classification, in which a three-tower PToP mapping is used to fuse their multiscale features. By introducing maximum correntropy criterion, Li \emph{et al.} \cite{HL2020} proposed a dual-channel robust capsule network (dual-channel CapsNet) for the fusion of HSI and LiDAR data. However, it should be noted that a fixed size of the convolution kernel is used for all classes, which may lead to the absence of multiscale information for different classes. In this case, the complementarity of HSI and LiDAR data is not fully exploited, since spectral and spatial information cannot be effectively integrated.

The attention mechanism is an important technique derived from computational neuroscience \cite{JL2013}. It allows a given model to automatically locate and capture the significant information from the input. Since Bahdanau \emph{et al.} \cite{DB2014} firstly utilized it to select reference words from source sentences, numerous works have demonstrated that, with the help of attention mechanisms, deep learning-based models can obtain better feature representation ability in many fields of computer vision \cite{AV2017}-\cite{JZ2019}. Several works have applied attention mechanisms to remote sensing problems. Cui \emph{et al.} \cite{ZC2019} proposed a dense attention pyramid network for ship detection in synthetic aperture radar (SAR) images, in which a convolutional block attention module (with spatial and channel-wise attention) is designed for highlighting salient features at specific scales. Chen \emph{et al.} \cite{JC2019} improved the faster region-based CNN by using multiscale (spatial) and channel-wise attention for object detection in remote sensing images. With a skip-connected encoder-decoder model, the work in \cite{CW2019} developed an end-to-end multiscale visual attention network for highlighting objects and suppressing background regions. Wang \emph{et al.} \cite{QW2019} proposed an end-to-end attention recurrent CNN for classification of very high-resolution (VHR) remote sensing scenes. Regarding HSI classification tasks, an attention-based inception model was designed in \cite{ZX2018} which can accurately model spatial information in HSI data. Mou \emph{et al.} \cite{LM2019} put forward a learnable spectral attention module (prior to CNN-based classification) for selecting informative bands. By combining an attention mechanism and RNNs, a spatial-spectral visual attention-driven feature extraction model was designed in \cite{JH2019}.

In this paper, a new dual-channel spatial, spectral and multiscale attention ConvLSTM neural network (dual-channel $A^{3}$CLNN) model is developed for the classification of multisource (i.e., HSI and LiDAR) remote sensing data. Specifically, three types of attention mechanisms are designed for extracting spectral- and spatial-enhanced multiscale features. Furthermore, a novel three-level fusion strategy is designed for effectively integrating the information coming from the HSI and LiDAR data. In the first-level fusion stage, composite attention learning is proposed for fully exploiting spatial and spectral information in the LiDAR and HSI data. Then, both types of features are cascaded as the input of the classification layer, which is the intermediate stage. Since the order of magnitude of HSI features is much larger than that of the LiDAR features, in the third-level fusion stage the LiDAR features are reused at the top of a fusion network to make full use of the LiDAR data source on the classification performance. To effectively train the proposed model, a stepwise training strategy is designed, in which these two branches (HSI and LiDAR) are first trained individually to obtain the primary features, and then --inspired by the idea of transfer learning \cite{JY2014}-- these features are used for initializing a fusion network that extracts high-level features. Finally, a multi-task loss function is designed to achieve a better optimization of the proposed dual-channel $A^{3}$CLNN model. The main contributions of this work can be summarized as follows.

\hangindent 2.5em
(1) Considering the wealth of spectral and spatial information presented in HSI and LiDAR data, we develop novel and learnable spectral and spatial attention modules to obtain spectral- and spatial-enhanced features.

\hangindent 2.5em
(2) For different classes, a fixed-scale feature extraction strategy may be inappropriate due to the different scale information contained in these classes. To solve this problem, a learnable multiscale residual attention module is further designed that enhances the multiscale information representation ability of the whole model.

\hangindent 2.5em
(3) A three-level fusion strategy is proposed. Particularly, a composite attention learning module that combines spectral and spatial attention is designed as a two-level attention strategy that makes better use of the spectral and spatial information. In the training stage, a stepwise training strategy (with a multi-task loss function) is designed for optimizing the proposed dual-channel $A^{3}$CLNN model, which can effectively accelerate its convergence speed.

The remainder of the paper is organized as follows. Section II reviews the ConvLSTM2D, ConvLSTM3D, and the attention mechanism. In Section III, the proposed dual-channel $A^{3}$CLNN model is described in detail. An exhaustive analysis of parameter settings and a quantitative evaluation of the proposed model are given in Section IV. Section V concludes the paper with some remarks and hints at plausible future research lines.

\begin{figure*}[htbp]
\centering
\setlength{\abovecaptionskip}{-8pt}
\begin{center}
\includegraphics[width=5.8in]{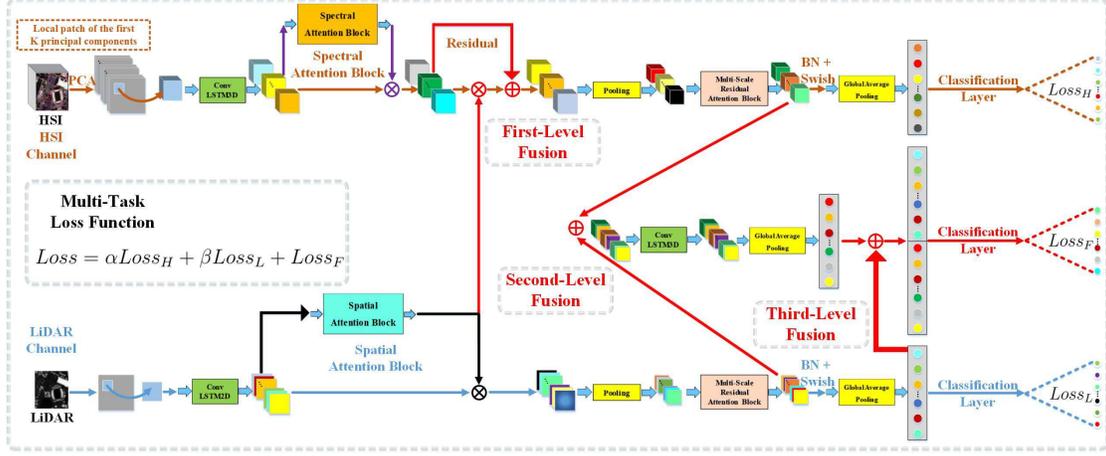}
\end{center}
\centering
\caption{Architecture of the proposed dual-channel $A^{3}$CLNN model.}
\vspace{-0.4cm}
\end{figure*}
\section{Related Work}
\subsection*{A. ConvLSTM2D and ConvLSTM3D}
As a modification and an extended version of LSTM, Shi \emph{et al.} \cite{XS2015} developed a ConvLSTM cell by extending the data processing method in LSTM to a 2-D convolution operation, with which plenty of the ConvLSTM-based deep models have been built for HSI classification \cite{HW2017}, \cite{AS2018}, \cite{MS2019}, \cite{QL2017}, \cite{WH2019}. Inspired by \cite{WH2019}, it is further named ConvLSTM2D cell for convenience. However, as shown in \cite{QL2017} and \cite{WH2019}, due to the special structure of the ConvLSTM2D cell, 3-D HSI data must be decomposed into a 2-D sequence when used as the input of the model, which may lose the intrinsic structure of HSI data.

To better preserve the intrinsic structure of HSI data, the ConvLSTM3D cell is further extended from the ConvLSTM2D cell, with which Hu \emph{et al.} \cite{WH2019} proposed a novel and effective SSCL3DNN model for HSI classification. Nevertheless, there is still much room to further improve the performance of SSCL3DNN. For example, the multiscale information is not considered, and the characteristics of the ConvLSTM3D layer for modeling long-term dependencies are not fully utilized.

\subsection*{B. Attention Mechanism}
After the attention mechanism was first introduced into deep learning in \cite{DB2014}, an increasing number of attention-driven deep learning-based models have been proposed. These models were able to improve the feature representation ability in many research fields. In \cite{AV2017}, Vaswani \emph{et al.} introduced the following equation to calculate the output of the attention mechanism:
\begin{IEEEeqnarray}{rCl}
Attention(Q, K, V) &=& softmax(f(Q,K))V,
\end{IEEEeqnarray}
where $f(\cdot)$ is the attention function, and $Q$, $K$, and $V$ are the inputs. $softmax(\cdot)$ denotes the softmax function used for normalization. Specifically, there are two common attention functions, i.e., additive attention \cite{DB2014} and dot-product attention \cite{AV2017}, where their corresponding definitions can be written as:
\begin{IEEEeqnarray}{rCl}
f_{additive}(Q, K) &=& W_{Q}Q + W_{K}K \notag \\
f_{dot-product}(Q, K) &=& QK^{T},
\end{IEEEeqnarray}
where $W_{Q}$ and $W_{K}$ are the parameter weights, and $T$ represents the transpose of the matrix.

The most common way for introducing an attention mechanism into deep learning is to build a hard part selection subnetwork or a soft mask branch \cite{HG2019}. By using residual learning, a residual attention module was built for soft pixel-level attention learning in \cite{FW2017}, and then applied to image classification. In addition, channel-wise attention \cite{JH2018}, spatial and temporal-wise attention \cite{QL2019}, spatial and channel-wise attention \cite{ZC2019}, and spatial-spectral attention \cite{JH2019} have also been proposed for feature enhancement. However, to the best of our knowledge, there have been no effective implementations of an attention mechanism for fusion and classification of multisource remote sensing data. In the following section, we develop a new composite attention learning module that combines spatial and spectral attention learning and a multiscale (residual) attention learning module for effectively combining HSI and LiDAR data.

\section{Dual-Channel $A^{3}$CLNN}
\subsection*{A. Architecture Overview}
It is well known that HSI data consist of many bands carrying plenty of spectral information, while LiDAR data are rich in height (spatial) information \cite{XX2018}. This motivates us to build a classification model upon a two-branch framework that fully exploits the complementary information from both sources of information.
\begin{figure}[!htbp]
\centering
\begin{center}
\includegraphics[width=2.9in]{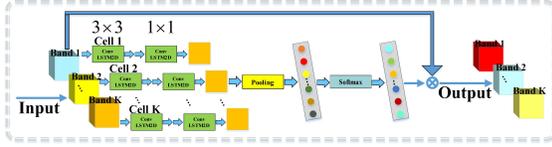}
\end{center}
\centering
\caption{Structure of the spectral attention block (SeAB).}
\vspace{-0.4cm}
\end{figure}
\begin{figure}[!htbp]
\centering
\begin{center}
\includegraphics[width=2.9in]{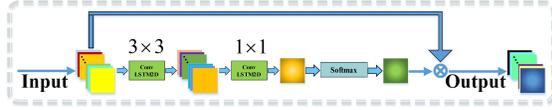}
\end{center}
\centering
\caption{Structure of the spatial attention block (SaAB).}
\vspace{-0.4cm}
\end{figure}
\begin{figure}[!htbp] 
\centering
\begin{center}
\includegraphics[width=3.0in]{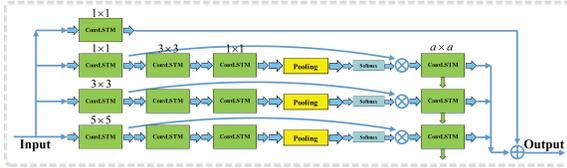}
\end{center}
\centering
\caption{Proposed multiscale residual attention block (MSRAB).}
\vspace{-0.4cm}
\end{figure}

The overall framework of the proposed dual-channel $A^{3}$CLNN model is graphically represented in Fig. 1. A spectral attention block (SeAB) and a spatial attention block (SaAB) are first proposed for composite attention learning (see subsection III-B). A multiscale residual attention block (MSRAB) is designed in subsection III-C. In subsection III-D and III-E, the HSI branch (marked with yellow arrows in Fig. 1) and the LiDAR branch (marked with blue arrows in Fig. 1) are described in detail. The proposed three-level fusion strategy (marked with red arrows in Fig. 1) is described in subsection III-F. Finally, subsection III-G describes the multi-task loss function and the stepwise training strategy.

\subsection*{B. Composite Attention Learning}
\subsubsection{Spectral Attention Block (SeAB)}
An effective and learnable SeAB module is designed to learn more discriminative and spectral-enhanced feature representation. The structure of this module is described in Fig. 2.

Let $X^{H}_{l}\in R^{w_{l} \times h_{l} \times s_{l} \times c_{l}}$ denote the output of the $l$th ConvLSTM3D layer (or the original HSI data), where $s_{l}$, $w_{l}$, $h_{l}$, and $c_{l}$ are respectively the number of spectral bands, width, height, and channel number. The purpose of SeAB is to learn an attention vector $\alpha^{H}_{Se}$.

As shown in Fig. 2, a $3\times3$ and a $1\times1$ ConvLSTM2D layers, a spatial pooling layer, and a softmax function comprise the main backbone of the proposed SeAB module. Firstly, $X^{H}_{l}$ is decomposed into $s_{l}$ 2-D components along the spectral dimension, and converted into a sequence, i.e., \(\left\{X^{H}_{l1}, \ldots, X^{H}_{lt}, \ldots, X^{H}_{ls_{l}} \right\}, t \in \left\{1,2,\ldots,s_{l} \right\}\), which are then fed (one by one) to the $3\times3$ ConvLSTM2D layer to model the long-range dependencies in the spectral domain. Another $1\times1$ ConvLSTM2D layer is added to generate an unnormalized attention map $Z^{H}_{Se}$, with size $ w_{l} \times h_{l} \times s_{l} \times1$. A spatial pooling operation is then applied to $Z^{H}_{Se}$ to transform it into an unnormalized attention vector $z^{H}_{Se}$, with length $s_{l}$. Finally, $z^{H}_{Se}$ is fed to a softmax function to yield the normalized attention vector $\alpha^{H}_{Se}$, which is multiplied by the input to yield a spectral-enhanced feature representation. The output $\hat{X^{H}_{l}}$ of SeAB can be described as:
\begin{IEEEeqnarray}{rCl}
z^{H}_{Se} &=& f_{p}(f_{CL2D1}(f_{CL2D3}(X^{H}_{l}))) \notag \\
\alpha^{H}_{Se} &=& softmax(z^{H}_{Se}) \notag \\
\hat{X^{H}_{l}} &=& X^{H}_{l} \odot \alpha^{H}_{Se},
\end{IEEEeqnarray}
where $\odot$ is an element-based product operation. $f_{CL2D3}(\cdot)$ and $f_{CL2D1}(\cdot)$ respectively indicate the $3\times3$ and $1\times1$ ConvLSTM2D layers. $f_{p}(\cdot)$ is a spatial pooling layer with size $w_{l} \times h_{l}$, and $softmax(\cdot)$ represents the softmax function. Particularly, the dimension time\_step in each ConvLSTM2D layer is set to $s_{l}$.

It should be noted that the input of SeAB can be either the original data or the output of the last layer. The special structure of SeAB makes it a feature enhancement module that can be added to any layer of the whole network to obtain spectral-enhanced feature representation. In our experiments, SeAB is added to the ConvLSTM3D layer.

\subsubsection{Spatial Attention Block (SaAB)}
Unlike HSI data, LiDAR data provide rich elevation information, which means information about the height and shape of targets \cite{XX2018}. We design an SaAB module for exploiting the spatial information in LiDAR data, which will lead to more effective spatial-enhanced feature representation.

Let $X^{L}_{l}\in R^{w_{l} \times h_{l} \times c_{l}}$ be the output of the $l$th ConvLSTM2D layer (or the original LiDAR data), in which $w_{l}$, $h_{l}$, and $c_{l}$ are defined in accordance with those in the SeAB. SaAB is constructed to learn an attention map $\alpha^{L}_{Sa}$.

The structure of the proposed SaAB module is shown in Fig. 3. The main backbone of this module is given by a $3\times3$ and a $1\times1$ ConvLSTM2D layer, and a softmax function. Firstly, the $3\times3$ and $1\times1$ ConvLSTM2D layers are used to generate an unnormalized attention map $Z^{L}_{Sa}$ with size $ w_{l} \times h_{l} \times 1$. After that, a softmax function is utilized to generate a normalized attention map $\alpha^{L}_{Sa}$. The forward propagation of SaAB can be written as:
\begin{IEEEeqnarray}{rCl}
Z^{L}_{Sa} &=& f_{CL2D1}(f_{CL2D3}(X^{L}_{l})) \notag \\
\alpha^{L}_{Sa} &=& softmax(Z^{L}_{Sa}) \notag \\
\hat{X^{L}_{l}} &=& X^{L}_{l} \odot \alpha^{L}_{Sa},
\end{IEEEeqnarray}
where $\hat{X^{L}_{l}}$ denotes the output of SaAB. The definitions of $\odot$, $f_{CL2D3}(\cdot)$, $f_{CL2D1}(\cdot)$, and $softmax(\cdot)$ are similar to that in (3). However, different from SeAB, the dimension time\_step in the SaAB module is fixed to 1.

Similar to SeAB, SaAB can be treated as an effective spatial feature extractor, and used by any layer of a deep learning-based model to obtain spatial-enhanced features. In our experiment, SaAB is utilized after the ConvLSTM2D layer in the LiDAR branch.

Based on the two aforementioned attention blocks, an effective composite attention learning approach is proposed for jointly learning the spatial-spectral features. Our approach can efficiently exploit spectral and spatial information, and enhance the feature extraction ability of the whole model. A more detailed description of the attention mechanism is given in subsection III-F.

\subsection*{C. Multiscale Residual Attention Block (MSRAB)}
In multisource remote sensing data, different classes may comprise different scale information, which means classification models using uniform scale to extract features may not meet the scaling requirements of different classes. Therefore, it is necessary to design a multiscale feature extractor able to properly describe multiscale information. An MSRAB module (integrating residual learning and attention mechanism) is proposed.

Taking the LiDAR branch as an example. Let $X^{L}_{l}\in R^{\tau_{l} \times w_{l} \times h_{l} \times c_{l}}$ and $X^{L}_{l+1}$ denote the input and output of the MSRAB module, where $\tau_{l}$ is the dimension time\_step of the ConvLSTM2D layer. The structure of the MSRAB module is given in Fig. 4. In particular, the first column is the multiscale feature extraction function realized by the ConvLSTM2D layers, which uses different fields of perception to capture multiscale information from the input (using different scales such as $1\times1$, $3\times3$, and $5\times5$). These features are cascaded in the time\_step dimension of the ConvLSTM2D layer. After the ConvLSTM2D layers, the obtained multiscale features are further learned in non-linear fashion, yielding an unnormalized attention map $Z^{L}_{MSR}\in R^{3\tau_{l} \times w_{l} \times h_{l} \times c_{l}}$. Then, after a global average pooling (GAP) layer, the unnormalized attention map $Z^{L}_{MSR}$ is converted to an unnormalized attention vector $z^{L}_{MSR}$ with length $3\tau_{l}$, which is further normalized by a softmax function to generate a multiscale attention vector $\alpha^{L}_{MSR}$. The output $O^{L}_{MSR}$ of the multiscale attention part in the MSRAB module is written as:
\begin{IEEEeqnarray}{rCl}
T &=& [f_{1}(X^{L}_{l}), f_{3}(X^{L}_{l}), f_{5}(X^{L}_{l})] \notag \\
z^{L}_{MSR} &=& f_{p}(f_{CL2D1}(f_{CL2D3}(T))) \notag \\
\alpha^{L}_{MSR} &=& softmax(z^{L}_{MSR}) \notag \\
O^{L}_{MSR} &=&  T \odot \alpha^{L}_{MSR},
\end{IEEEeqnarray}
where $[\cdot, \cdot, \cdot]$ denotes the concatenation operation, and $f_{1}(\cdot)$, $f_{3}(\cdot)$, and $f_{5}(\cdot)$ are the multiscale feature extraction functions realized by the ConvLSTM2D layer.

Unlike the cascading approach in \cite{CS2015} and \cite{YT2019}, the output of MSRAB is fed to a fusion layer (built by a ConvLSTM2D layer) to model long-term dependencies in the multiscale dimension. Furthermore, residual learning is also applied to mitigate the gradient vanishing or explosion problems through a feature reuse mechanism. The forward propagation of MSRAB is expressed as:
\begin{IEEEeqnarray}{rCl}
X^{L}_{l+1} &=& f_{CL2Da}(O^{L}_{MSR})+ X^{L}_{l},
\end{IEEEeqnarray}
where $f_{CL2Da}(\cdot)$ denotes an $a \times a$ ConvLSTM2D layer.

Specially, the structure of MSRAB in the HSI branch is similar to that in the LiDAR branch. However, the dimensions $X^{H}_{l}$ and $X^{H}_{l+1}$ need to be extended to $ R^{\tau_{l} \times w_{l} \times h_{l} \times s_{l} \times c_{l}}$ and $ R^{3\tau_{l+1} \times w_{l+1} \times h_{l+1} \times s_{l+1} \times c_{l+1}}$, in which $s_{l}$ and $s_{l+1}$ are the spectral dimensions. $f_{1}(\cdot)$, $f_{3}(\cdot)$, $f_{5}(\cdot)$, $f_{CL2D1}(\cdot)$, $f_{CL2D3}(\cdot)$, and $f_{CL2Da}(\cdot)$ in (5) and (6) are implemented by ConvLSTM3D layers.

Similar to SeAB and SaAB, MSRAB can be used as a multiscale information enhancement module to bring a larger receptive field to the whole model, and MSRAB can adaptively focus on important areas at each scale. 

\subsection*{D. Multiscale Spectral Attention Neural Network (MSSeA) for the HSI Branch}
An MSSeA model is proposed for HSI branch, which is marked in Fig. 1 by yellow arrows. The backbone of MSSeA consists of a ConvLSTM3D layer, an SeAB module, a pooling layer, an MSRAB model, a GAP layer, and a classification layer. Specifically, considering the redundant information presented in the original HSI data, principal component analysis (PCA) is used for spectral dimensionality reduction in our experiments.

Let $W \times H \times D$ denote the size of the original HSI data, where $W$, $H$, and $D$ are the width, height, and the number of the spectral bands, respectively. In the data preprocessing stage, the first $K$ principal components are retained, and the pixels in a local neighborhood window with size $s \times s$ are extracted to account for the spatial-contextual information around each pixel $x$. Accordingly, the whole data associated to pixel $x$ can be represented as $X^{H}\in R^{s \times s \times K}$, which is also the input of MSSeA. To make the data meet the format requirements of the ConvLSTM3D layer, $X^{H}$ is decomposed into $\tau$ 3-D components and then converted into a sequence with length $\tau$, as indicated below:
\begin{IEEEeqnarray}{rCl}
X^{H} \Rightarrow \left\{X^{H}_1, \ldots, X^{H}_t, \ldots, X^{H}_{\tau} \right\},
\end{IEEEeqnarray}
where $X^{H}_t$ is the \(t\)th 3-D component, and $t \in \left\{1,2,\ldots, \tau \right\}$. $\tau$ is the dimension time\_step in the ConvLSTM3D layer (fixed here to 1).

Then, this sequence is fed into $l$ cascaded ConvLSTM3D layers (one by one) to extract shallow spatial-spectral features. To facilitate subsequent variable representation, the output of the $l$th ConvLSTM3D layer is written as $X^{H}_{l}\in R^{\tau_{l} \times w_{l} \times h_{l} \times s_{l} \times c_{l}}$, where $\tau_{l}$ is the dimension time\_step, and the size of the convolution kernel is $k^{H}_{l} \times k^{H}_{l} \times k^{H}_{l}$. Following each ConvLSTM3D layer, SeAB is applied to extract the spectral-enhanced features and, according to (3), the enhanced features can be expressed as $\hat{X^{H}_{l}}$. In our experiments, $l$ is set to 1.

Furthermore, to measure the multiscale information and meet the scale requirements of different classes, the spectral-enhanced features $\hat{X^{H}_{l}}$, obtained after a pooling layer, are input to an MSRAB module to learn the multiscale information. From (6), the extracted multiscale features can be written as $X^{H}_{l+1} \in R^{3\tau_{l+1} \times w_{l+1} \times h_{l+1} \times s_{l+1} \times c_{l+1}}$. Inspired by \cite{PR2017} and for the sake of accelerating convergence and solving the gradient vanishing problem, a batch normalization (BN) layer and a swish function are used for regularization.

Then, we apply a GAP layer \cite{ML2014} at the top of MSSeA --instead of the fully connected (FC) layer-- to map the feature space to class label space, which can directly endow each channel with the actual category meaning, regularize the whole model, and prevent over-fitting to some degree. In addition, this strategy can effectively solve the problem of having too many parameters in the FC layer, relaxing the limitations imposed to the model by the resolution of the input. The forward propagation of the GAP layer in MSSeA can be expressed as:
\begin{IEEEeqnarray}{rCl}
X^{H}_{GAP} &=& f_{GAP}(X^{H}_{l+1}),
\end{IEEEeqnarray}
where $f_{GAP}(\cdot)$ and $X^{H}_{GAP} \in R^{1 \times c_{l+1}}$ denote the expression and output of the GAP layer, respectively.

Finally, a classification layer (with the softmax function) follows the GAP layer to predict the conditional probability distribution $P^{H}_{c}=P(y=c|X^{H}_{GAP},W,b)=\frac{e^{(W_{c}X^{H}_{GAP}+b_c)}}{\sum_{j=1}^Ne^{(W_{j}X^{H}_{GAP}+b_j)}}$ of each class $c$, where $c \in {1,2,\ldots,N}$, and $N$ denotes the number of classes.

To obtain the final classification results, the cross entropy is selected as the loss function to optimize the HSI branch, which is named $Loss_{H}$ for convenience.
\subsection*{E. Multiscale Spatial Attention Neural Network (MSSaA) for the LiDAR Branch}
Similar to subsection III-D, an MSSaA model is designed for the LiDAR branch (marked in Fig. 1 with blue arrows). A ConvLSTM2D layer, an SaAB module, a down-sample layer, an MSRAB module, a GAP layer, and a classification layer represent the backbone of it.

Let us assume that the size of the original LiDAR data is $W \times H$. In the data preparation stage, a $s \times s$ local spatial-contextual window around each pixel $x$ is first extracted, which is fed into the MSSaA and expressed as $X^{L}\in R^{s \times s}$. Due to the special structure of the ConvLSTM2D layer, $X^{L}$ needs to be decomposed into a sequence with $\tau$ 2-D components as follows:
\begin{IEEEeqnarray}{rCl}
X^{L} \Rightarrow \left\{X^{L}_1, \ldots, X^{L}_t, \ldots, X^{L}_{\tau} \right\},
\end{IEEEeqnarray}
where $X^{L}_t$ is the \(t\)th 2-D component and $t \in \left\{1,2,\ldots, \tau \right\}$. $\tau$ is the dimension time\_step in the ConvLSTM2D layer, which is set to 1 in our experiments.

Then, this sequence is fed into $l$ cascaded ConvLSTM2D layers (one by one) to extract the shallow spatial features. For convenience, the output of the $l$th ConvLSTM2D layer is described as $X^{L}_{l}\in R^{\tau_{l} \times w_{l} \times h_{l} \times c_{l}}$. After that, an SaAB module is applied to enhance the spatial information of the output of each ConvLSTM2D layer. According to (4), the enhanced features are written as $\hat{X^{L}_{l}}$. Furthermore, a pooling layer and an MSRAB are utilized for learning the multiscale information, and then the extracted multiscale features $X^{L}_{l+1} \in R^{3\tau_{l+1} \times w_{l+1} \times h_{l+1} \times c_{l+1}}$ are mapped into the category space by a GAP layer whose output is given by $X^{L}_{GAP}$. Particularly, $l$ is set to 1 in our experiments.

Finally, $X^{L}_{GAP}$ is fed to a softmax function to obtain the conditional probability distribution $P^{L}_{c}$, and the cross entropy is also set as the loss function $Loss_{L}$ of the LiDAR channel to yield the final classification results.


\subsection*{F. Three-Level Fusion Strategy}
An effective fusion network (with a three-level fusion strategy) is designed for making full use of the complementarity of the HSI and LiDAR data. This network is marked in Fig. 1 with red arrows.

In the first-level fusion, to fully exploit the (more complete) spatial information carried out by the LiDAR data to enhance the feature representation in the HSI branch, the spatial attention in SaAB is applied to the output of SeAB for composite attention learning. Furthermore, residual learning is also utilized to avoid the degradation problem. The forward propagation of this part is expressed as:
\begin{equation}
F_{\hat{X^{H}_{l}}} = \alpha^{L}_{Sa} \odot \hat{X^{H}_{l}} + \hat{X^{H}_{l}},
\end{equation}
where $F_{\hat{X^{H}_{l}}}$ denotes the spatial-spectral features of the HSI channel.

In the second-level fusion, the outputs of MSRAB in each branch are cascaded in the spectral dimension, and then, a $1\times1$ ConvLSTM3D layer and a GAP layer are utilized to fuse the cascaded features, as shown in Fig. 1. The forward propagation of this part is expressed as:
\begin{IEEEeqnarray}{rCl}
X^{F}_{GAP} = f_{GAP}(f_{CL3D1}([X^{H}_{l+1},X^{L}_{l+1}])),
\end{IEEEeqnarray}
where $X^{F}_{GAP}$ denotes the output of the GAP layer, and $f_{CL3D1}(\cdot)$ denotes the $1\times1$ ConvLSTM3D layer.

Finally, due to the fact that the order of magnitude of the HSI features is much larger than that of the LiDAR features, the impact of the LiDAR channel on the classification performance may need to be upscaled. Hence, at the top of the designed fusion network, the LiDAR features $X^{L}_{GAP}$ are reused by cascading them with the features in (11), i.e., the third-level fusion. The outputs of this part are written as $X^{F} = [X^{F}_{GAP}, X^{L}_{GAP}]$, and then fed into a softmax function to yield the probability distribution $P^{F}_{c}$. Similar to the HSI and LiDAR branches, the cross entropy is still used as the loss function $Loss_{F}$ to optimize the fusion network.

\subsection*{G. Loss Function and Network Training Strategy}

Given the loss functions in subsections III-D, III-E and III-F, a multi-task loss function for optimizing the proposed dual-channel $A^{3}$CLNN model is designed as:
\begin{equation}
Loss = \alpha Loss_{H} + \beta Loss_{L} + \gamma Loss_{F},
\end{equation}
where $\alpha$, $\beta$, and $\gamma$ are the scalar weights. For convenience, they are fixed to 1 in our experiments.

\begin{figure*}[htbp]
\centering
\setlength{\abovecaptionskip}{-8pt}
\begin{center}
\includegraphics[width=4.2in]{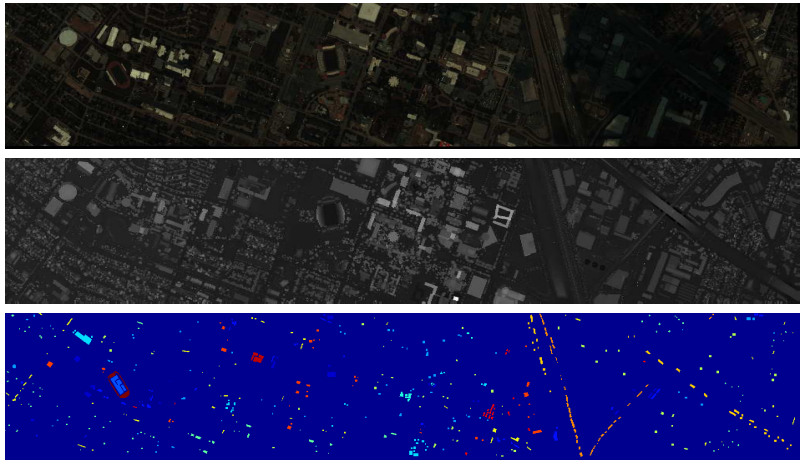}
\end{center}
\centering
\caption{(First row) false-color map (using bands 57, 27 and 17) of the Houston HSI data. (Second row) grayscale representation of the Houston LiDAR data. (Third row) ground-truth map of the Houston data set.}
\vspace{-0.6cm}
\end{figure*}
All the weights and biases in the proposed dual-channel $A^{3}$CLNN model need to be learned. In order to train the whole model adequately --different from the training strategies in \cite{XX2018}-- a novel and effective stepwise training approach is proposed. In the first stage, the LiDAR and HSI branches are trained using $N_{step1}$ and $N_{step2}$ epochs, respectively, which can provide the primary spatial features and spatial-spectral features to the designed fusion network and be regarded as pre-trained channels to replace the ones obtained by traditional random initialization. Then, inspired by transfer learning \cite{JY2014}, the proposed fusion network is initialized by these two pre-trained branches, and the multi-task loss function in (12) is further optimized in $N_{steps}$ epochs to yield the final classification results of the proposed model. This also accelerates the convergence speed of the whole model. The detailed training strategy is summarized in Algorithm 1.

\begin{algorithm}[H]
\footnotesize
\caption{Training Dual-Channel $A^{3}$CLNN for Multisource Remote Sensing Data Classification} 
\hspace*{0.02in} {\bf Input:} 
\hspace*{0.02in} {HSI data $X^{H}$; The LiDAR data $X^{L}$;}
\hspace*{0.02in} {Ground truth $Y$}\\
\hspace*{0.02in} {\bf Output:} 
\hspace*{0.001in} {Classification map $\Omega$}
\begin{algorithmic}[1]
\State Parameter setting and weights initialization
\While{step $\leq$ $N_{step1}$}\\ 
\hspace*{0.08in} Train the LiDAR branch by optimizing the loss function $Loss_{L}$\\
\hspace*{0.08in} Save model as the pre-trained LiDAR model
\EndWhile
\While{step $\leq$ $N_{step2}$} 
\hspace*{0.08in} Train the HSI branch by optimizing the loss function $Loss_{H}$\\
\hspace*{0.08in} Save model as the pre-trained HSI model
\EndWhile
\While{step $\leq$ $N_{steps}$} 
\hspace*{0.08in} Restore these two pre-trained models to initialize the fusion network\\
\hspace*{0.08in} Train the whole classification model by optimizing the multi-task loss function \ $Loss$
\EndWhile
\State \Return Classification map $\Omega$
\end{algorithmic}
\end{algorithm}

It should be pointed out that the adaptive momentum (ADAM) algorithm is adopted to optimize the three loss functions $Loss_{H}$, $Loss_{L}$, and $Loss$ with different learning rates, which are represented by $lr_{H}$, $lr_{L}$, and $lr$, respectively. Additional explanations on parameter settings will be given in Section IV.

\section{Experimental Results}
In order to quantitatively and qualitatively evaluate the performance of the proposed dual-channel $A^{3}$CLNN model, some state-of-the-art methods are selected for comparison, such as ELM \cite{WL2015}, SVM \cite{XX2018}, SSCL3DNN and SaCL2DNN \cite{WH2019}, two-branch CNN \cite{XX2018}, and dual-channel CapsNet \cite{HL2020}. The overall (OA), average accuracy (AA), and Kappa coefficient ($\kappa$) are utilized as the quantitative metrics to measure the classification performance of all algorithms. For the sake of eliminating the bias caused by random initialization of parameters in deep learning-based models, all experiments are repeated 10 times, and the average value is given for each quantitative metric. All our experiments have been conducted on a desktop PC with an Intel Core i7-8700 processor and an Nvidia GeForce GTX 1080ti GPU.

\subsection*{A. Experimental Data}
In our experiments, two HSI + LiDAR data sets, i.e., Houston data set and Trento data set, are considered to evaluate the performance of our dual-channel $A^{3}$CLNN. According to \cite{BR2017} and \cite{MK2015}, the false-color maps, grayscale representations, ground-truth maps, and the training samples for each data set are respectively presented in Figs. 5-6 and Tables I-II. In the following, we describe these data sets in more detail:
\begin{figure}[H]
\centering
\begin{center}
\includegraphics[width=2.6in]{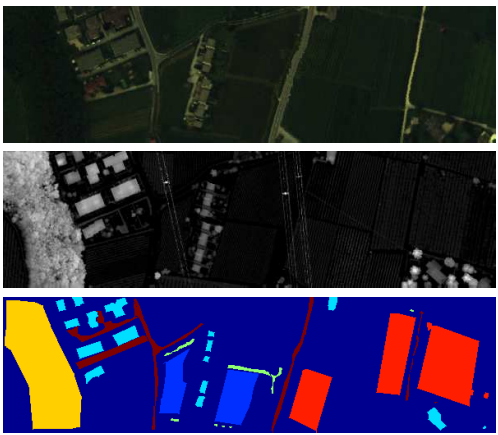}
\end{center}
\centering
\caption{(First row) false-color map (using bands 25, 15 and 2) of the Trento HSI data. (Second row) grayscale representation of the Trento LiDAR data. (Third row) ground-truth map of the Trento data set.}
\end{figure}

\emph{1) Houston Data:} These data were captured in 2012 by the Compact Airborne Spectrographic Imager (CASI) sensor over the University of Houston campus and the surrounding area. The data was introduced in the 2013 IEEE Geoscience and Remote Sensing Society (GRSS) Data Fusion contest. Its size is $349\times1905$ pixels, with spatial resolution of 2.5 m. There are 144 bands in the wavelength range from 0.38 to 1.05 $\mu$m and 15 distinguishable class labels. These data, including the reference classes, are available online from the IEEE GRSS Data and Algorithm Standard Evaluation Website: \url{http://dase.grss-ieee.org/}.
\begin{table}[H]
    \centering
    \renewcommand\thetable{\Roman{table}}
    \renewcommand\tabcolsep{3.0pt}
    \caption{Number of Training Samples for The Houston Data}
    \scriptsize
    \begin{tabular}{p{7cm}<{\centering}}
        \specialrule{0.1em}{0.5pt}{0.5pt}
    	$ \begin{tabular}{p{0.5cm}<{\centering}p{0.7cm}<{\centering}p{2.2cm}<{\centering}p{1cm}<{\centering}p{0.8cm}<{\centering}}
            \specialrule{0em}{1pt}{1pt}
            NO.  & Color & Class & Training & Total\\ \specialrule{0em}{0pt}{1pt}
            \end{tabular}$ \\\specialrule{0.05em}{0pt}{0.5pt}
        $\begin{tabular}{p{7cm}<{\centering}}
            \begin{tabular}{p{0.8cm}<{\centering}p{0.4cm}<{\centering}p{2.6cm}<{\centering}p{0.5cm}<{\centering}p{1.4cm}<{\centering}}
            1 & \multicolumn{1}{>{\columncolor{mycH1}}l}{} & \textup{Health grass} & 198 & 1251\\
            2 & \multicolumn{1}{>{\columncolor{mycH2}}l}{} & \textup{Stressed grass} & 190 & 1254\\
            3 & \multicolumn{1}{>{\columncolor{mycH3}}l}{} & \textup{Synthetic grass} & 192 & 697\\
            4 & \multicolumn{1}{>{\columncolor{mycH4}}l}{} & \textup{Tress} & 188 & 1244\\
            5 & \multicolumn{1}{>{\columncolor{mycH5}}l}{} & \textup{Soil} & 186 & 1242\\
            6 & \multicolumn{1}{>{\columncolor{mycH6}}l}{} & \textup{Water} & 182 & 325\\
            7 & \multicolumn{1}{>{\columncolor{mycH7}}l}{} & \textup{Residential} & 196 & 1268\\
            8 & \multicolumn{1}{>{\columncolor{mycH8}}l}{} & \textup{Commercial} & 191 & 1244\\
            9 & \multicolumn{1}{>{\columncolor{mycH9}}l}{} & \textup{Road} & 193 & 1252\\
            10& \multicolumn{1}{>{\columncolor{mycH10}}l}{} & \textup{Highway} & 191 & 1227\\
            11& \multicolumn{1}{>{\columncolor{mycH11}}l}{} & \textup{Railway} & 181 & 1235\\
            12& \multicolumn{1}{>{\columncolor{mycH12}}l}{} & \textup{Parking lot 1} & 192 & 1233\\
            13& \multicolumn{1}{>{\columncolor{mycH13}}l}{} & \textup{Parking lot 2} & 184 & 469\\
            14& \multicolumn{1}{>{\columncolor{mycH14}}l}{} & \textup{Tennis court} & 181 & 428\\
            15& \multicolumn{1}{>{\columncolor{mycH15}}l}{} & \textup{Running track} & 187 & 660 \\
              &  &  &  & \\
            \end{tabular}
        \end{tabular}$\\\specialrule{0.05em}{1pt}{1pt}
    	$ \begin{tabular}{p{0.5cm}<{\centering}p{0.7cm}<{\centering}p{2.2cm}<{\centering}p{1cm}<{\centering}p{0.8cm}<{\centering}}
            Total&   &  & 2832 & 15029\\
        \end{tabular}$ \\
        \specialrule{0.1em}{0.5pt}{0.5pt}
    \end{tabular}
\end{table}
\begin{table}[H]
    \centering
    \renewcommand\thetable{\Roman{table}}
    \renewcommand\tabcolsep{3.0pt}
    \caption{Number of Training Samples for The Trento Data}
    \scriptsize
    \begin{tabular}{p{7cm}<{\centering}}
        \specialrule{0.1em}{0.5pt}{0.5pt}
    	$ \begin{tabular}{p{0.5cm}<{\centering}p{0.7cm}<{\centering}p{2.2cm}<{\centering}p{1cm}<{\centering}p{0.8cm}<{\centering}}
            \specialrule{0em}{1pt}{1pt}
            NO.  & Color & Class & Training & Total\\ \specialrule{0em}{0pt}{1pt}
            \end{tabular}$ \\\specialrule{0.05em}{0pt}{0.5pt}
        $\begin{tabular}{p{7cm}<{\centering}}
            \begin{tabular}{p{0.8cm}<{\centering}p{0.4cm}<{\centering}p{2.6cm}<{\centering}p{0.5cm}<{\centering}p{1.4cm}<{\centering}}
            1 & \multicolumn{1}{>{\columncolor{mycT1}}l}{} & \textup{Apple trees} & 129 & 4034\\
            2 & \multicolumn{1}{>{\columncolor{mycT2}}l}{} & \textup{Buildings} & 125 & 2903\\
            3 & \multicolumn{1}{>{\columncolor{mycT3}}l}{} & \textup{Ground} & 105 & 479\\
            4 & \multicolumn{1}{>{\columncolor{mycT4}}l}{} & \textup{Woods} & 154 & 9123\\
            5 & \multicolumn{1}{>{\columncolor{mycT5}}l}{} & \textup{Vineyard} & 184 & 10501\\
            6 & \multicolumn{1}{>{\columncolor{mycT6}}l}{} & \textup{Roads} & 122 & 3374\\
              &  &  &  & \\
            \end{tabular}
        \end{tabular}$\\\specialrule{0.05em}{1pt}{1pt}
    	$ \begin{tabular}{p{0.5cm}<{\centering}p{0.7cm}<{\centering}p{2.2cm}<{\centering}p{1cm}<{\centering}p{0.8cm}<{\centering}}
            Total &   &   & 819  & 30414\\
            \end{tabular}$\\
        \specialrule{0.1em}{0.5pt}{0.5pt}
    \end{tabular}
\vspace{-0.4cm}
\end{table}

\emph{2) Trento Data:} These data were collected by the AISA Eagle sensor over a rural area in Trento, Italy. The data comprises $600\times166$ pixels with spatial resolution of 1 m, 6 ground-truth classes, and 63 bands in the spectral range from 420.89 to 989.09 nm.

\begin{figure}[H]
    \centering
    \renewcommand\tabcolsep{1.0pt}
    \vspace{-0.2cm}
    \scriptsize 
    \begin{tabular}{p{4.1cm}<{\centering}p{4cm}<{\centering}}
        \begin{minipage}[t]{1\linewidth}
            \centering
            \includegraphics[width=1.45in]{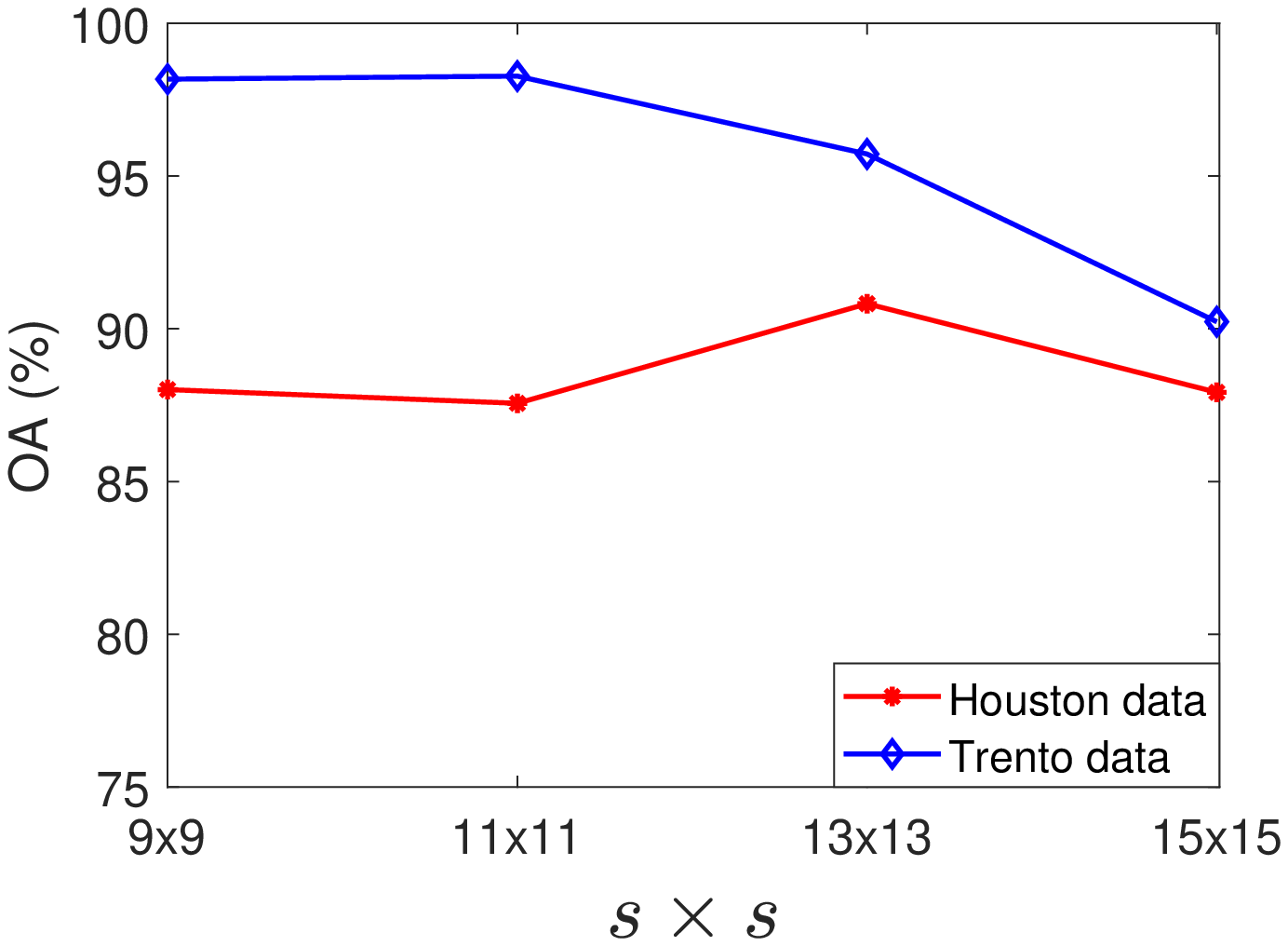}
        \end{minipage}
        & \begin{minipage}[t]{1\linewidth}
            \centering
            \includegraphics[width=1.45in]{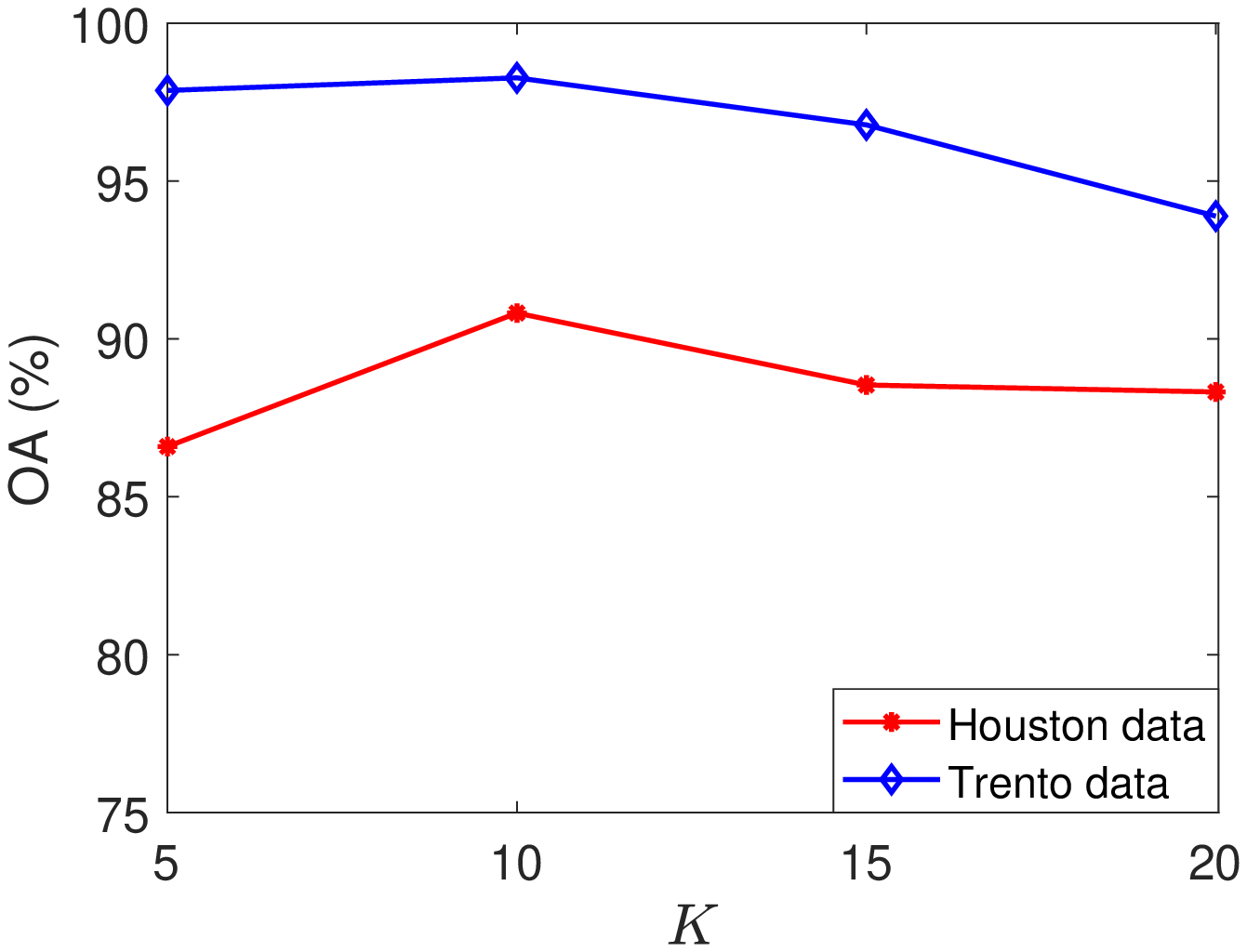}
        \end{minipage}\\
        (a) & (b) \\
    \end{tabular}
    \begin{tabular}{p{4cm}<{\centering}p{4cm}<{\centering}}
        \begin{minipage}[t]{1\linewidth}
            \centering
            \includegraphics[width=1.45in]{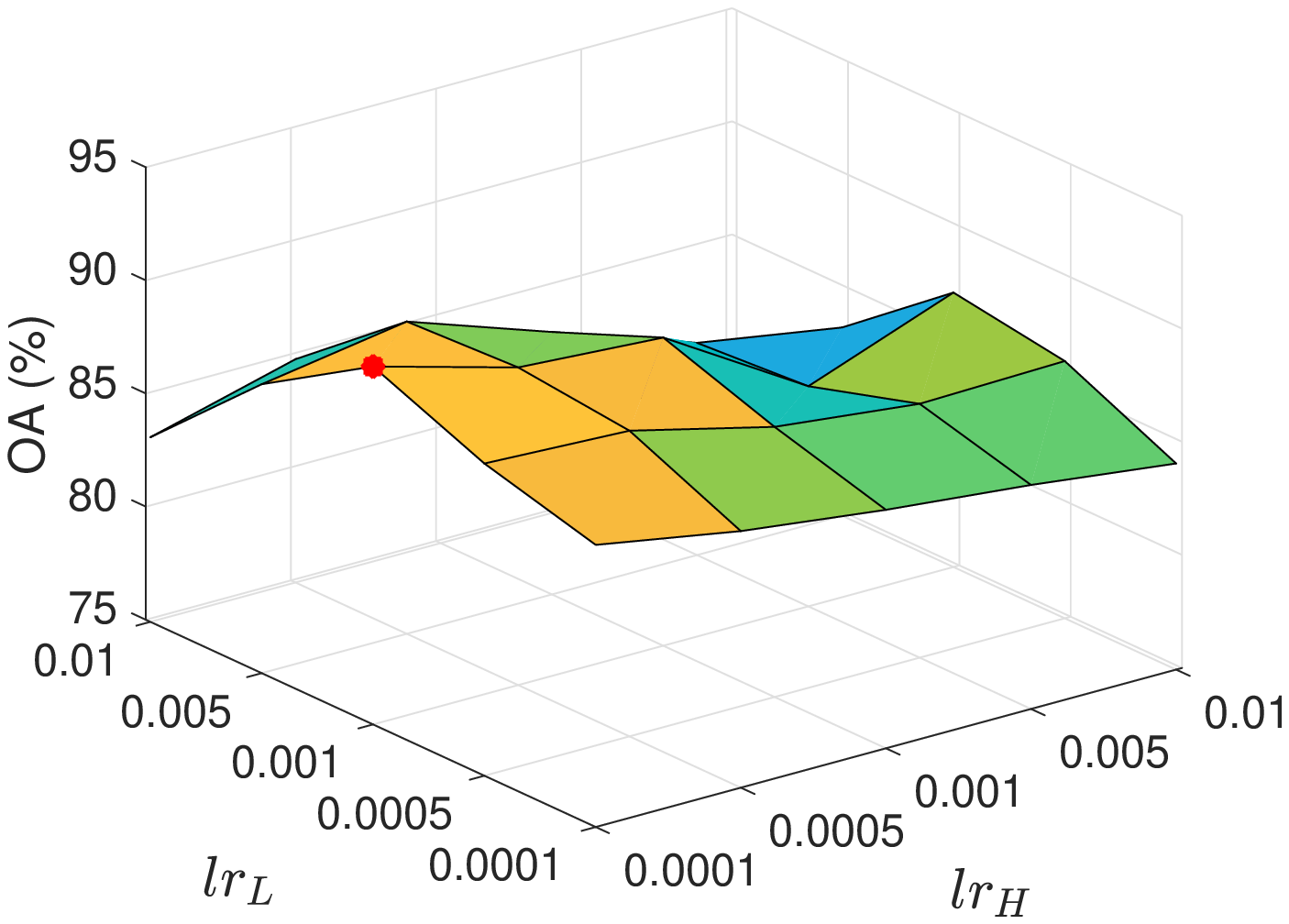}
        \end{minipage}
        & \begin{minipage}[t]{1\linewidth}
            \centering
            \includegraphics[width=1.45in]{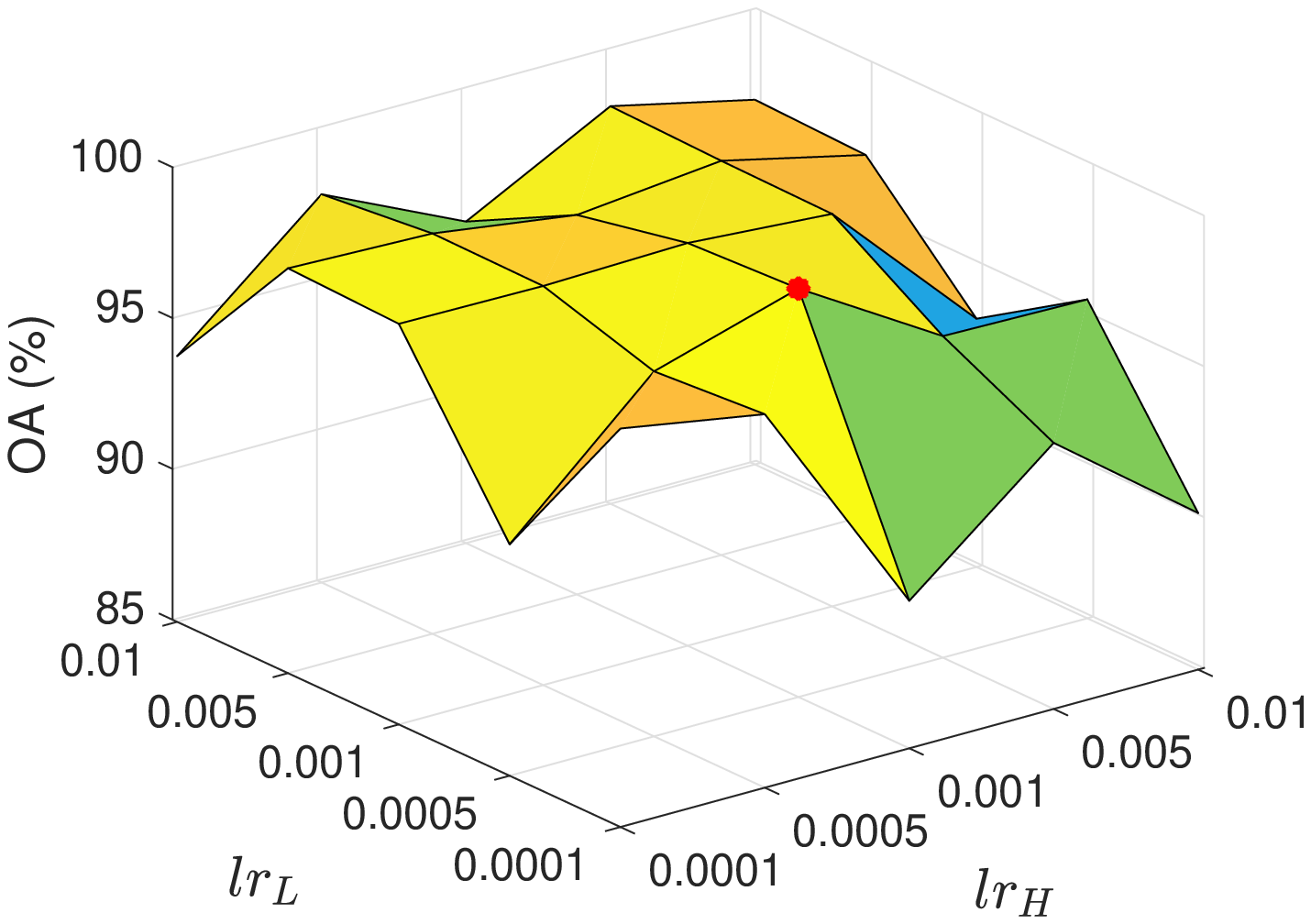}
        \end{minipage}\\
        (c) & (d) \\
    \end{tabular}
    \begin{tabular}{p{4cm}<{\centering}p{4cm}<{\centering}}
        \begin{minipage}[t]{1\linewidth}
            \centering
            \includegraphics[width=1.45in]{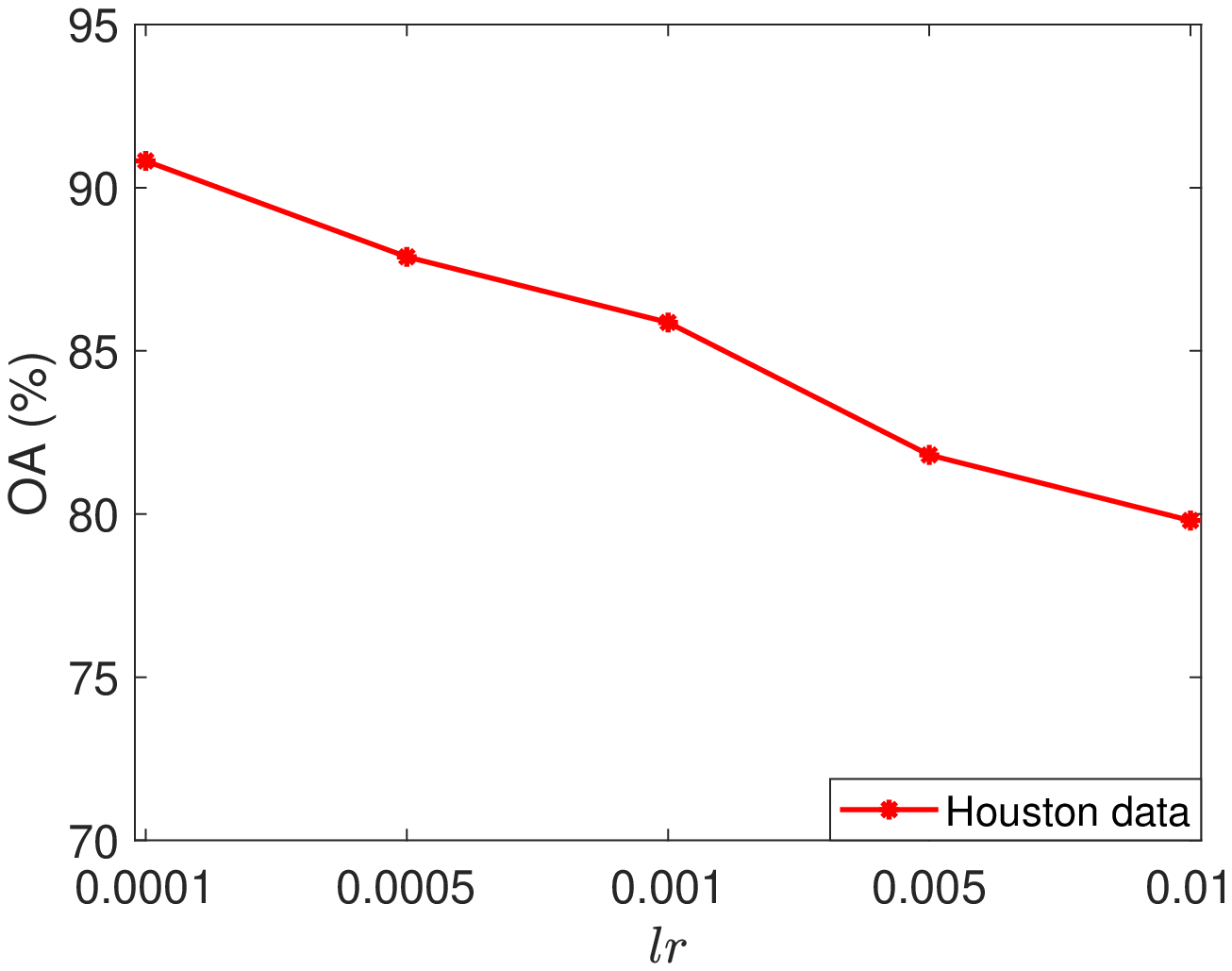}
        \end{minipage}
        & \begin{minipage}[t]{1\linewidth}
            \centering
            \includegraphics[width=1.5in]{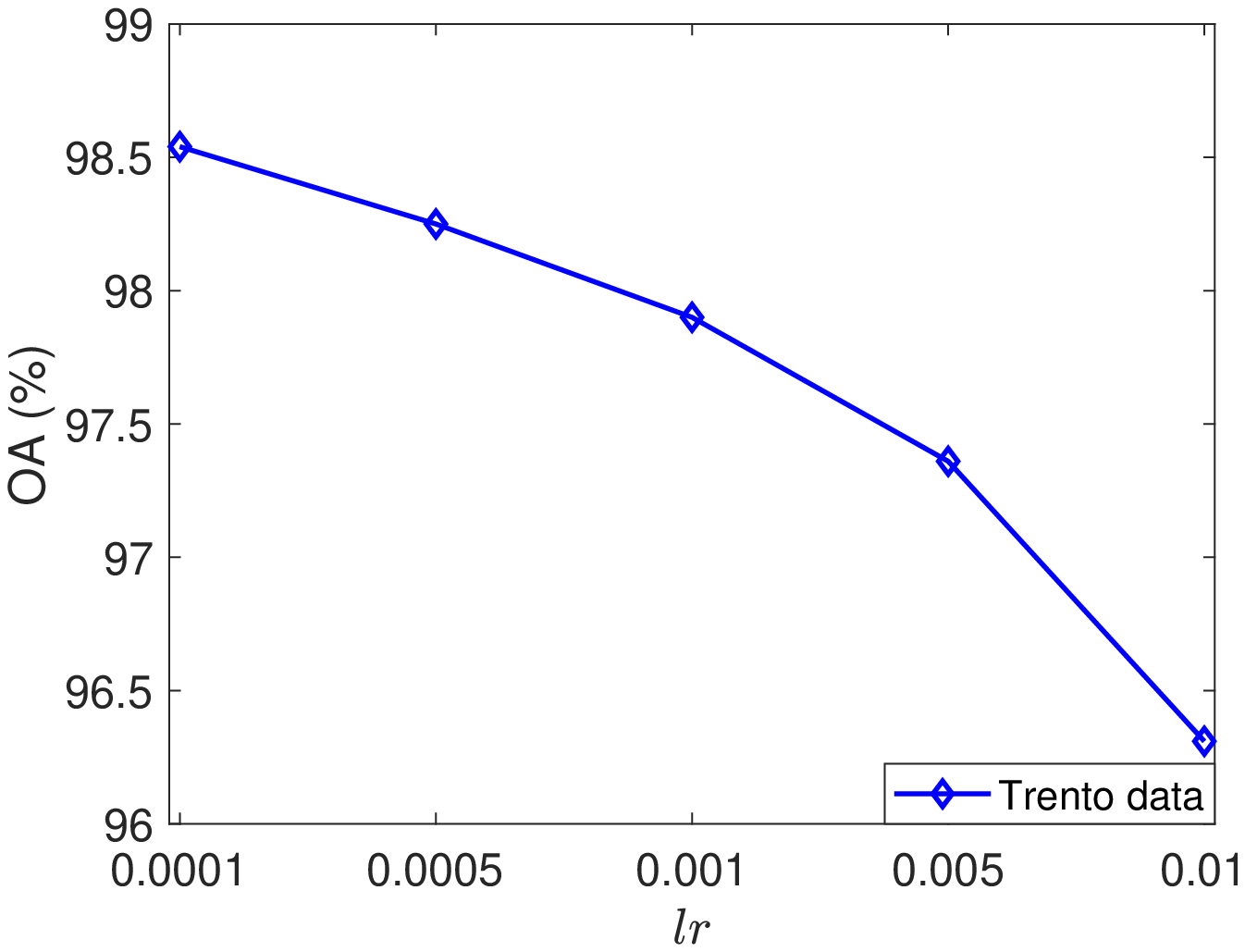}
        \end{minipage}\\
        (e) & (f) \\
    \end{tabular}
    \begin{tabular}{p{4cm}<{\centering}p{4cm}<{\centering}}
        \begin{minipage}[t]{1\linewidth}
            \centering
            \includegraphics[width=1.5in]{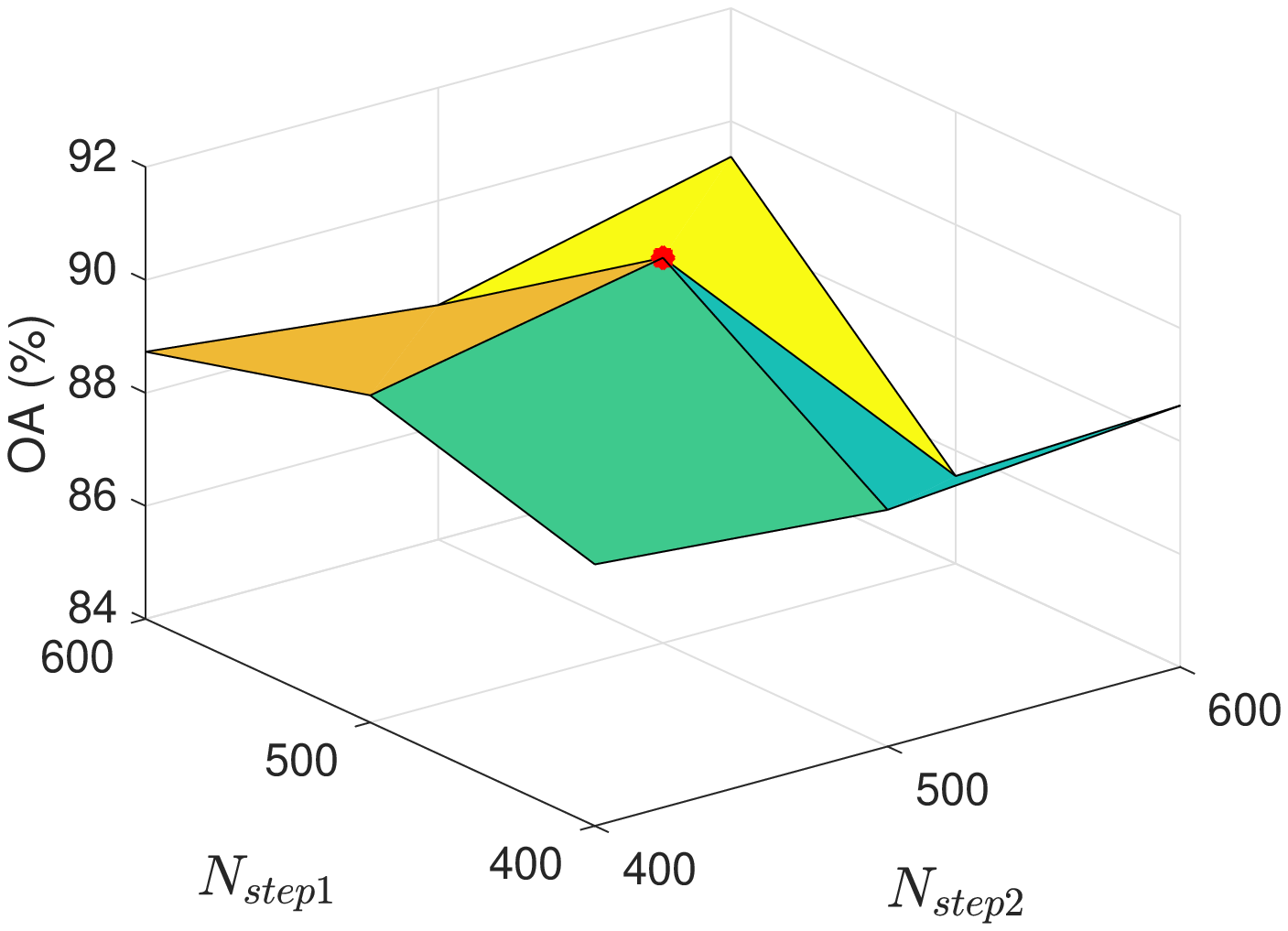}
        \end{minipage}
        & \begin{minipage}[t]{1\linewidth}
            \centering
            \includegraphics[width=1.5in]{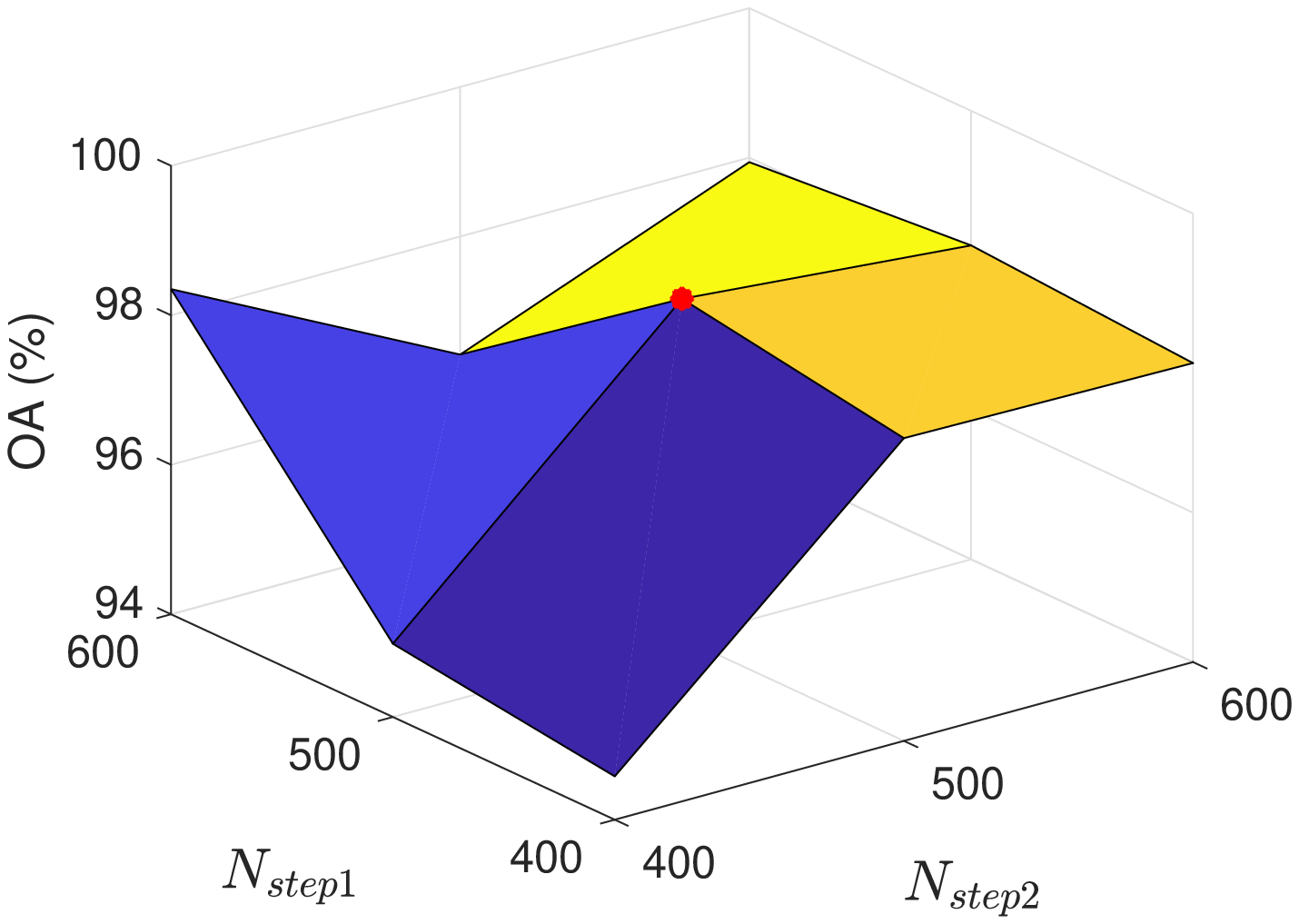}
        \end{minipage}\\
        (g) & (h) \\
    \end{tabular}
\caption{Overall accuracy (\%) achieved by the proposed dual-channel $A^{3}$CLNN model with different parameters for the Houston and Trento data sets. (a) Size $s \times s$ of the local spatial window. (b) Number $K$ of principal components. (c)-(d) Learning rates $lr_{H}$ and $lr_{L}$. (e)-(f) Learning rate $lr$. (g)-(h) Training epochs $N_{step1}$ and $N_{step2}$.}
\vspace{-0.2cm}
\end{figure}
\subsection*{B. Parameter Settings}
As in \cite{YC2016} and \cite{WH2019}, PCA is also utilized as the dimension reduction approach. For the HSI branch in the dual-channel $A^{3}$CLNN model, the first $K$ components after PCA are retained as the spectral information.

For the compared algorithms, the parameter settings of SVM, ELM, SSCL3DNN, SaCL2DNN, two-branch CNN, and dual-channel CapsNet are obtained according to \cite{XX2018}, \cite{WL2015}, \cite{WH2019}, \cite{XX2018}, \cite{HL2020} to achieve quasi-optimal performance. For the proposed dual-channel $A^{3}$CLNN model in Fig. 1, there are some parameters that need to be tuned, i.e., the size ($s \times s$) of the local spatial window, the number ($K$) of the principal components, the size ($k \times k$) of the convolution kernels, the number ($M$) of feature maps, the value of the dropout operation, the learning rates ($lr_{H}$, $lr_{L}$, and $lr$), and the training epochs ($N_{step1}$ and $N_{step2}$). At first, $K$ is fixed to 10, and the value of the dropout operation is 0.5. The learning rates [$lr_{H}$, $lr_{L}$, $lr$] for the Houston data set are set to [0.0001, 0.001, 0.0001], respectively, and to [0.001, 0.0005, 0.0001] for the Trento data set, respectively. The training epochs $N_{step1}$ and $N_{step2}$ for the two data sets are fixed to 500. Parameter $M$ in the first layer of each branch, the $a \times a$ ConvLSTM layer of MSRAB, and the ConvLSTM3D layer of the fusion network is respectively fixed to $\left\{{32,64,128}\right\}$. The number $N_{steps}$ of training epochs is fixed to 1200 for the two data sets. After that, $s$ is searched from $\left\{{9, 11, 13, 15}\right\}$ for the two data sets, and $k$ is set to a value in the range $\left\{{3, 4, 5}\right\}$. Based on the above parameter settings, the experimental results for analyzing the effect of using different spatial window sizes on the classification performance are reported in Fig. 7(a), from which it is obvious that the optimal size of the local window is set to $13\times13$ (Houston data) and $11\times11$ (Trento data), respectively.

Furthermore, an optimal number $K$ is generated from a given set $\left\{{5, 10, 15, 20}\right\}$, and the OA achieved by the proposed method (for different values of parameter $K$) is shown in Fig. 7(b), from which it can be seen that the optimal value of $K$ is 10 for the two data sets.

Then, the performance on different values of $M$ for four different combinations: $\left\{{8,16,32}\right\}$, $\left\{{16,32,64}\right\}$, $\left\{{32,64,128}\right\}$, and $\left\{{64,128,256}\right\}$ is analyzed in Table III. From Table III, we can infer that the optimal number of feature maps is $\left\{{32,64,128}\right\}$ (Houston data) and $\left\{{16,32,64}\right\}$ (Trento data), respectively.

To reduce the occurrence of overfitting problem, the dropout operation is used as a training trick, and the experiments for analyzing the influence of its different values on the classification performance are conducted. From Table IV, the optimal value of the dropout operation is set to 0.5 for the two data sets.

The learning rate is one of the parameters that controls the convergence rate of the proposed dual-channel $A^{3}$CLNN model in the training process, and the optimal values of $lr_{H}$, $lr_{L}$ and $lr$ are searched in the range $\left\{{0.0001,0.0005,0.001,0.005,0.01}\right\}$. Concretely, Figs. 7(c)-(d) report the experimental results obtained after tuning the learning rates when $lr$ was fixed to 0.0001. From Figs. 7(c)-(d), it is evident that the optimal value of [$lr_{H}$, $lr_{L}$] for the two data sets is [0.0001, 0.001] (Houston data) and [0.001, 0.0005] (Trento data), respectively. Furthermore, the experimental results obtained after tuning $lr$ are shown in Figs. 7(e)-(f), from which it can be seen that the learning rate $lr$ for the two data sets can be fixed to 0.0001.
\begin{table}[H]
    \centering
    \renewcommand\thetable{\Roman{table}}
    \renewcommand\tabcolsep{3.0pt}
    \caption{Sensitivity Comparison and Analysis of The Feature Maps Obtained for Different Values of $M$}
    \scriptsize
    \begin{tabular}{c|c|c}
        \specialrule{0.1em}{0.5pt}{0.5pt}
        $M$ \quad& Houston Data Set \quad& Trento Data Set\\
        \hline
        $\left\{{8 ,16,32}\right\}$  & 85.96 &97.56\\
        $\left\{{16,32,64}\right\}$  & 87.87 &\textbf{98.54}\\
        $\left\{{32,64,128}\right\}$ & \textbf{90.82} &98.27\\
        $\left\{{64,128,256}\right\}$& 88.64 &96.35\\
        \specialrule{0.1em}{0.5pt}{0.5pt}
      \end{tabular}
\vspace{-0.4cm}
\end{table}
\begin{table}[H]
    \centering
    \renewcommand\thetable{\Roman{table}}
    \renewcommand\tabcolsep{3.0pt}
    \vspace{-0.2cm}
    \caption{Sensitivity Comparison and Analysis of Different Values of The Dropout Operation}
    \scriptsize
    \begin{tabular}{c|c|c}
        \specialrule{0.1em}{0.5pt}{0.5pt}
        Dropout \quad& Houston Data Set \quad& Trento Data Set\\
        \hline
        $0.4$ & 89.42          &98.22\\
        $0.5$ & \textbf{90.82} &\textbf{98.54}\\
        $0.6$ & 88.69          &98.14\\
        \specialrule{0.1em}{0.5pt}{0.5pt}
      \end{tabular}
\vspace{-0.2cm}
\end{table}

\begin{table}[H]
    \centering
    \renewcommand\tabcolsep{1.0pt}
    \caption{Parameter Settings for The Houston Data Set}
    \scriptsize
    \begin{tabular}{p{7.95cm}<{\centering}}
        \specialrule{0.1em}{0.5pt}{0pt}
    	$ \begin{tabular}{c|c| p{2.2cm}| p{1.7cm}}
        \multirow{2}*{Layer Name}  \quad& \multirow{2}*{Kernel Size} \quad& Output Size for HSI Channel \quad& Output Size for LiDAR Channel\\
        \hline
        Input              &                         & \multicolumn{1}{c|}{$13\times13\times10\times1$}& \multicolumn{1}{c}{$13\times13\times1$}\\
        ConvLSTM3D         & $3\times3\times3$       & $13\times13\times10\times32$                    \\
        ConvLSTM2D         & $3\times3$              &                                            & \multicolumn{1}{c}{$13\times13\times32$}\\
        SeAB               &                         & $13\times13\times10\times32$                    \\
        SaAB               &                         &                                            & \multicolumn{1}{c}{$13\times13\times32$}\\
        \specialrule{0.1em}{0pt}{0pt}
        First-Level Fusion &                         & \multicolumn{2}{c}{$13\times13\times10\times32$}\\
        \specialrule{0.1em}{0pt}{0pt}
        MaxPooling3D       & $2\times2\times2$       & \multicolumn{1}{c|}{$7\times7\times5\times32$}  \\
        MaxPooling2D       & $2\times2 $             &                                            & \multicolumn{1}{c}{$7\times7\times32$}\\
        MSRAB(3D)          & $4\times4\times4$        & $3\times7\times7\times5\times64$ \\
        MSRAB(2D)          & $3\times3$               &                                          &\multicolumn{1}{c}{$3\times7\times7\times64$}\\
        \specialrule{0.1em}{0pt}{0pt}
        Second-Level Fusion& $1\times1\times1$       & \multicolumn{2}{c}{$3\times7\times7\times6\times128$}\\
        Dropout            & 0.5                     \\
        GAP3D              & $3\times7\times7\times6$& \multicolumn{2}{c}{$1\times1\times1\times1\times128$}\\
        \specialrule{0.1em}{0pt}{0pt}
        Dropout            & 0.5                     &                    & \\
        GAP3D              & $3\times7\times7\times5$& $1\times1\times1\times1\times64$                \\
        GAP2D              & $3\times7\times7$       &                                           &\multicolumn{1}{c}{$1\times1\times1\times64$}\\
        \specialrule{0.1em}{0pt}{0pt}
        Third-Level Fusion &                         & \multicolumn{2}{c}{$1\times1\times1\times1\times192$}\\
        Dropout            & 0.5                     \\
        \specialrule{0.1em}{0pt}{0pt}
        Softmax            &                         & \multicolumn{1}{c|}{13}                   & \multicolumn{1}{c}{13}\\
    \end{tabular}$ \\
    \specialrule{0.1em}{0.5pt}{0pt}
    \end{tabular}
\vspace{-0.4cm}
\end{table}
\begin{table}[H]
    \centering
    \vspace{-0.2cm}  
    \renewcommand\tabcolsep{1.0pt}
    \caption{Parameter Settings for The Trento Data Set}
    \scriptsize
    \begin{tabular}{p{7.95cm}<{\centering}}
        \specialrule{0.1em}{0.5pt}{0pt}
    	$ \begin{tabular}{c|c| p{2.2cm}| p{1.7cm}}
        \multirow{2}*{Layer Name}  \quad& \multirow{2}*{Kernel Size} \quad& Output Size for HSI Channel \quad& Output Size for LiDAR Channel\\
        \hline
        Input              &                         & \multicolumn{1}{c|}{$11\times11\times10\times1$}& \multicolumn{1}{c}{$11\times11\times2$}\\
        ConvLSTM3D         & $3\times3\times3$       & $11\times11\times10\times16$                    \\
        ConvLSTM2D         & $3\times3$              &                                            & \multicolumn{1}{c}{$11\times11\times16$}\\
        SeAB               &                         & $11\times11\times10\times16$                    \\
        SaAB               &                         &                                            & \multicolumn{1}{c}{$11\times11\times16$}\\
        \specialrule{0.1em}{0pt}{0pt}
        First-Level Fusion &                         & \multicolumn{2}{c}{$11\times11\times10\times16$}\\
        \specialrule{0.1em}{0pt}{0pt}
        MaxPooling3D       & $2\times2\times2$       & \multicolumn{1}{c|}{$6\times6\times5\times16$}  \\
        MaxPooling2D       & $2\times2 $             &                                            & \multicolumn{1}{c}{$6\times6\times16$}\\
        MSRAB(3D)          & $4\times4\times4$        & $3\times6\times6\times5\times32$ \\
        MSRAB(2D)          & $3\times3$               &                                          &\multicolumn{1}{c}{$3\times6\times6\times32$}\\
        \specialrule{0.1em}{0pt}{0pt}
        Second-Level Fusion& $1\times1\times1$       & \multicolumn{2}{c}{$3\times6\times6\times6\times64$}\\
        Dropout            & 0.5                     \\
        GAP3D              & $3\times6\times6\times6$& \multicolumn{2}{c}{$1\times1\times1\times1\times64$}\\
        \specialrule{0.1em}{0pt}{0pt}
        Dropout            & 0.5                     &                    & \\
        GAP3D              & $3\times6\times6\times5$& $1\times1\times1\times1\times32$                \\
        GAP2D              & $3\times6\times6$       &                                           &\multicolumn{1}{c}{$1\times1\times1\times32$}\\
        \specialrule{0.1em}{0pt}{0pt}
        Third-Level Fusion &                         & \multicolumn{2}{c}{$1\times1\times1\times1\times96$}\\
        Dropout            & 0.5                     \\
        \specialrule{0.1em}{0pt}{0pt}
        Softmax            &                         & \multicolumn{1}{c|}{13}                   & \multicolumn{1}{c}{13}\\
    \end{tabular}$ \\
    \specialrule{0.1em}{0.5pt}{0pt}
    \end{tabular}
\end{table}

Finally, for the training epochs $N_{step1}$ and $N_{step2}$, we carry out the experiments to study the effect of the training epochs of these two pre-training channels, and the optimal $N_{step1}$ and $N_{step2}$ are selected from $\left\{{400,500,600}\right\}$. As shown in Figs. 7(g)-(h), the optimal [$N_{step1}$, $N_{step2}$] for the two data sets is [500, 500]. The detailed parameter settings for the proposed model for the two data sets are reported in Tables V-VI.

\subsection*{C. Classification Performance}
According to \cite{BR2017} and \cite{MK2015}, we evaluate the performance of the considered classification algorithms on the Houston and Trento data sets, using the available training samples. The obtained results are reported in Tables I-II. Note that the data enhancement technology in \cite{XX2018} is utilized to extend training sets for two-branch CNN and dual-channel CapsNet.
\begin{table}[H]
    \centering
    \renewcommand\thetable{\Roman{table}}
    \renewcommand\tabcolsep{3.0pt}
    \caption{Classification Performance of Each Branch: HSI (H) AND LIDAR (L)}
    \scriptsize
    \begin{tabular}{p{3.1cm}<{\centering}p{1.6cm}<{\centering}p{2.0cm}<{\centering}}
    \specialrule{0.1em}{0.5pt}{0.5pt}
      $ \begin{tabular}{c|cc|}
        \multicolumn{1}{c}{\multirow{2}*{Data Set}} \quad& \multicolumn{2}{|c|}{Proposed (L)}\\
        \quad& \multicolumn{1}{c}{OA} \quad& \multicolumn{1}{c|}{Kappa}\\
        \hline
        Houston Data  & 59.83&56.48\\
        Trento Data   & 89.31&85.98\\
      \end{tabular}$ &
      $ \begin{tabular}{c c c|}
        \multicolumn{3}{c|}{Proposed (H)}\\
        \quad& OA \quad& Kappa\\
        \hline
        & 87.00          & 85.90          \\
        & 97.65          & 96.86          \\
      \end{tabular}$ &
      $ \begin{tabular}{c c c}
        \multicolumn{3}{c}{Proposed (H+L)}\\
        \quad& OA \quad& Kappa\\
        \hline
        & 90.55          & 89.75         \\
        & 98.73          & 98.31          \\
      \end{tabular}$ \\
    \specialrule{0.1em}{0.5pt}{0.5pt}
    \end{tabular}
\vspace{-0.4cm}
\end{table}

\begin{table*}
    \centering
    \renewcommand\thetable{\Roman{table}}
    \renewcommand\tabcolsep{3.0pt}
    \caption{Classification Results Achieved by Different Approaches for The Houston Data Set}
    \scriptsize
    \begin{tabular}{p{0.44cm}|c c c c c c c c c c c c} 
        \specialrule{0.1em}{0.5pt}{0.5pt}
        \multirow{2}*{Class} \quad& SVM \quad& SVM \quad& ELM \quad& ELM \quad& Two-Branch \quad& Two-Branch \quad& Dual-Channel \quad& Dual-Channel \quad& SSCL3DNN \quad& SSCL3DNN \quad& Proposed \quad& Proposed\\
        &\multicolumn{1}{c}{(H)}&\multicolumn{1}{c}{(H+L)}&\multicolumn{1}{c}{(H)}&\multicolumn{1}{c}{(H+L)}\quad&\multicolumn{1}{c}{CNN(H)}
        &\multicolumn{1}{c}{CNN(H+L)}&\multicolumn{1}{c}{CapsNet(H)}&\multicolumn{1}{c}{CapsNet(H+L)}
        &\multicolumn{1}{c}{(H)}&\multicolumn{1}{c}{(Merge, H+L)}&\multicolumn{1}{c}{(H)}&\multicolumn{1}{c}{(H+L)}\\
        \hline
        1 & 81.01 & 82.53& 82.15& 82.24& \multicolumn{1}{c}{94.85} & \multicolumn{1}{c}{\textbf{97.98}} &80.63&81.39& \multicolumn{1}{c}{81.04}& \multicolumn{1}{c}{82.05}&79.84& 81.73\\
        2 & 82.24 & 84.77& 82.93& 82.99& \multicolumn{1}{c}{82.59} & \multicolumn{1}{c}{\textbf{89.44}} &81.95&83.08& \multicolumn{1}{c}{84.21}& \multicolumn{1}{c}{80.98}&85.15& 84.43\\
        3 & 82.97 & 86.93& 95.45& 96.96& \multicolumn{1}{c}{30.32} & \multicolumn{1}{c}{53.20} &94.46&\textbf{97.43}& \multicolumn{1}{c}{72.21}& \multicolumn{1}{c}{89.44}&93.20& 91.49\\
        4 & 90.81 & 95.83& 90.44& 91.41& \multicolumn{1}{c}{\textbf{98.30}} & \multicolumn{1}{c}{96.73} &90.06&88.64& \multicolumn{1}{c}{90.79}& \multicolumn{1}{c}{90.85}&89.20& 96.72\\
        5 & 97.63 & 97.54& 99.59& 98.96& \multicolumn{1}{c}{95.61} & \multicolumn{1}{c}{96.88}          &\textbf{100.00}&\textbf{100.00}& \multicolumn{1}{c}{\textbf{100.00}}& \multicolumn{1}{c}{99.78}&99.84& 99.97\\
        6 & 79.72 & 88.81& 71.79& 77.86& \multicolumn{1}{c}{76.41} & \multicolumn{1}{c}{24.31}          &89.51&95.10& \multicolumn{1}{c}{84.38}& \multicolumn{1}{c}{87.18}&95.34& \textbf{97.90}\\
        7 & 76.12 & 81.16& 80.32& 74.91& \multicolumn{1}{c}{\textbf{96.18}} & \multicolumn{1}{c}{87.47} &81.72&91.23& \multicolumn{1}{c}{89.55}& \multicolumn{1}{c}{91.51}&84.58& 87.06\\
        8 & 43.40 & 44.92& 64.10& 63.60& \multicolumn{1}{c}{70.54} & \multicolumn{1}{c}{89.02}          &71.51&92.40& \multicolumn{1}{c}{68.66}& \multicolumn{1}{c}{93.32}&81.83& \textbf{96.93}\\
        9 & 79.41 & 86.40& 72.62& 77.34& \multicolumn{1}{c}{82.31} & \multicolumn{1}{c}{86.85}          &72.24&80.64& \multicolumn{1}{c}{\textbf{87.91}}& \multicolumn{1}{c}{78.88}&86.02& 87.88\\
        10 & \textbf{90.15}& 59.75& 80.79& 58.91& \multicolumn{1}{c}{65.41} & \multicolumn{1}{c}{78.20} &62.26&65.54& \multicolumn{1}{c}{52.38}& \multicolumn{1}{c}{55.60}&60.42& 70.82\\
        11 & 63.00& 71.82& 68.22& 88.58& \multicolumn{1}{c}{78.69} & \multicolumn{1}{c}{90.88}          &72.87&88.99& \multicolumn{1}{c}{73.62}& \multicolumn{1}{c}{90.83}&95.70& \textbf{98.13}\\
        12 & 84.15& 92.41& 72.81& 78.87& \multicolumn{1}{c}{83.75} & \multicolumn{1}{c}{65.99}          &86.55&87.42& \multicolumn{1}{c}{93.05}& \multicolumn{1}{c}{91.80}&93.05& \textbf{94.65}\\
        13 & 89.82& 85.96& 42.57& 54.97& \multicolumn{1}{c}{94.25} & \multicolumn{1}{c}{\textbf{100.00}}&77.89&62.46&\multicolumn{1}{c}{92.05}& \multicolumn{1}{c}{85.96}&91.46& 96.02\\
        14 & 80.97& 83.00& 90.01& 92.44& \multicolumn{1}{c}{95.02} & \multicolumn{1}{c}{98.28} &93.93&95.95& \multicolumn{1}{c}{92.31}& \multicolumn{1}{c}{78.41}&\textbf{99.60}& 97.30\\
        15 & 66.60& 74.21& 84.00& 93.94& \multicolumn{1}{c}{91.10} & \multicolumn{1}{c}{96.37}          &94.93&96.41& \multicolumn{1}{c}{94.43}& \multicolumn{1}{c}{94.86}&\textbf{99.86}& 96.05\\
        \hline
        OA & 78.79         & 80.15& 79.52& 80.76& \multicolumn{1}{c}{77.79}& \multicolumn{1}{c}{83.15}  &81.53&86.61& \multicolumn{1}{c}{82.72}& \multicolumn{1}{c}{86.01}& 87.00& \textbf{90.55}\\
        AA & 79.20         & 81.07& 78.52& 80.96& \multicolumn{1}{c}{82.35}& \multicolumn{1}{c}{83.44}  &83.37&87.11& \multicolumn{1}{c}{83.79}& \multicolumn{1}{c}{86.10}& 89.01& \textbf{91.81}\\
        \(\kappa\) & 77.15 & 78.58& 77.74& 79.10& \multicolumn{1}{c}{75.95}& \multicolumn{1}{c}{81.73}  &80.01&85.50& \multicolumn{1}{c}{81.33}& \multicolumn{1}{c}{84.84}& 85.90& \textbf{89.75}\\
        \specialrule{0.1em}{0.5pt}{0.5pt}
    \end{tabular}
\vspace{-0.4cm}  
\end{table*}
\begin{table*}
    \centering
    \renewcommand\thetable{\Roman{table}}
    \renewcommand\tabcolsep{3.0pt}
    \caption{Classification Results Achieved by Different Approaches for The Trento Data Set}
    \scriptsize
    \begin{tabular}{p{0.44cm}|c c c c c c c c c c c c} 
        \specialrule{0.1em}{0.5pt}{0.5pt}
        \multirow{2}*{Class} \quad& SVM \quad& SVM \quad& ELM \quad& ELM \quad& Two-Branch \quad& Two-Branch \quad& Dual-Channel \quad& Dual-Channel \quad& SSCL3DNN \quad& SSCL3DNN \quad& Proposed \quad& Proposed\\
        &\multicolumn{1}{c}{(H)}&\multicolumn{1}{c}{(H+L)}&\multicolumn{1}{c}{(H)}&\multicolumn{1}{c}{(H+L)}\quad&\multicolumn{1}{c}{CNN(H)}
        &\multicolumn{1}{c}{CNN(H+L)}&\multicolumn{1}{c}{CapsNet(H)}&\multicolumn{1}{c}{CapsNet(H+L)}
        &\multicolumn{1}{c}{(H)}&\multicolumn{1}{c}{(Merge, H+L)}&\multicolumn{1}{c}{(H)}&\multicolumn{1}{c}{(H+L)}\\
        \hline
        1 & 59.59& 97.69& 89.31& 93.17   & \multicolumn{1}{c}{90.38}& \multicolumn{1}{c}{97.44} &98.46&97.15& \multicolumn{1}{c}{96.48}& \multicolumn{1}{c}{98.32}& 98.17& \textbf{98.92}\\
        2 & 34.55& 87.36& 71.55& 87.95   & \multicolumn{1}{c}{96.66}& \multicolumn{1}{c}{93.29} &94.14&99.07& \multicolumn{1}{c}{91.62}& \multicolumn{1}{c}{96.88}& 95.97& \textbf{99.14}\\
        3 & 92.69& 87.06& 92.21& 73.56   & \multicolumn{1}{c}{88.24}& \multicolumn{1}{c}{72.86} &91.44&97.29& \multicolumn{1}{c}{90.81}& \multicolumn{1}{c}{82.19}& 91.79& \textbf{98.12}\\
        4 & 98.61& 99.80& 97.58& 97.30   & \multicolumn{1}{c}{99.32}& \multicolumn{1}{c}{98.01} &96.17&\textbf{100.00}& \multicolumn{1}{c}{97.15}& \multicolumn{1}{c}{99.81}& 99.59& \textbf{100.00}\\
        5 & 97.93& 93.36& 87.57& 93.17   & \multicolumn{1}{c}{98.24}& \multicolumn{1}{c}{98.44} &98.93&94.62& \multicolumn{1}{c}{99.87}& \multicolumn{1}{c}{96.74}& 99.89& \textbf{99.95}\\
        6 & 79.30& 69.34& 59.84& 66.95   & \multicolumn{1}{c}{60.09}& \multicolumn{1}{c}{80.14} &72.21&\textbf{91.71}& \multicolumn{1}{c}{79.31}& \multicolumn{1}{c}{85.34}& 86.40& 90.57\\
        \hline
        OA & 84.89         & 92.69& 86.45& 90.85   & \multicolumn{1}{c}{93.20}& \multicolumn{1}{c}{95.36} &94.65&96.75& \multicolumn{1}{c}{95.50}& \multicolumn{1}{c}{96.46}& 97.65& \textbf{98.73}\\
        AA & 77.11         & 89.10& 83.01& 85.35   & \multicolumn{1}{c}{88.82}& \multicolumn{1}{c}{90.03} &91.89&96.64& \multicolumn{1}{c}{92.54}& \multicolumn{1}{c}{93.21}& 95.30& \textbf{97.78}\\
        \(\kappa\) & 79.45 & 90.22& 82.01& 87.81   & \multicolumn{1}{c}{90.92}& \multicolumn{1}{c}{93.81} &92.87&95.69& \multicolumn{1}{c}{93.99}& \multicolumn{1}{c}{95.30}& 96.86& \textbf{98.31}\\
        \specialrule{0.1em}{0.5pt}{0.5pt}
    \end{tabular}
\vspace{-0.4cm}  
\end{table*}
\begin{figure*}[htbp]
\centering
\setlength{\abovecaptionskip}{-8pt}
\begin{center}
\includegraphics[width=5.8in]{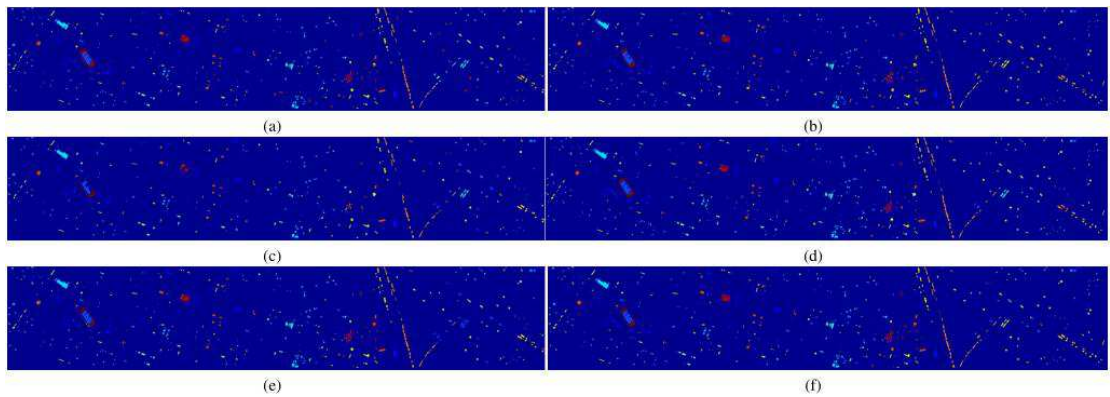}
\end{center}
\centering
\caption{Classification maps obtained by different approaches for the Houston data set. (a) SVM (80.15\%). (b) ELM (80.76\%). (c) Two-branch CNN (83.15\%). (d) dual-channel CapsNet (86.61\%). (e) SSCL3DNN (86.01\%). (f) The proposed dual-channel $A^{3}$CLNN (90.55\%).}
\vspace{-0.2cm}
\end{figure*} 
\begin{figure*}[htbp]
\centering
\setlength{\abovecaptionskip}{-8pt}
\begin{center}
\includegraphics[width=5.5in]{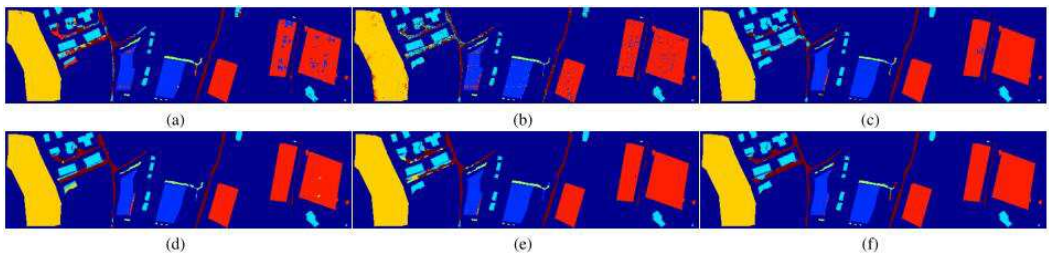}
\end{center}
\centering
\caption{Classification maps obtained by different approaches for the Trento data set. (a) SVM (92.69\%). (b) ELM (90.85\%). (c) Two-branch CNN (95.36\%). (d) dual-channel CapsNet (96.75\%). (e) SSCL3DNN (96.46\%). (f) The proposed dual-channel $A^{3}$CLNN (98.73\%).}
\vspace{-0.2cm}
\end{figure*}

\begin{table*}
    \centering
    \renewcommand\thetable{\Roman{table}}
    \renewcommand\tabcolsep{3.0pt}
    \caption{Classification Results Obtained by Different Approaches for The LiDAR Data of The Two Considered Data Sets}
    \scriptsize
    \begin{tabular}{c|c c c c c c |c c c c c c}
        \specialrule{0.1em}{0.5pt}{0.5pt}
        \multirow{3}*{Class} & \multicolumn{6}{c|}{Houston Data Set} & \multicolumn{6}{c}{Trento Data Set}\\
        \quad& SVM \quad& ELM \quad& Two-Branch \quad& Dual-Channel \quad& SaCL2DNN \quad& Proposed  \quad& SVM \quad& ELM \quad& Two-Branch \quad& Dual-Channel \quad& SaCL2DNN \quad& Proposed\\
        &\multicolumn{1}{c}{(L)}&\multicolumn{1}{c}{(L)}&\multicolumn{1}{c}{CNN(L)}&\multicolumn{1}{c}{CapsNet(L)}&\multicolumn{1}{c}{(L)}
        &\multicolumn{1}{c|}{(L)}&\multicolumn{1}{c}{(L)}&\multicolumn{1}{c}{(L)}&\multicolumn{1}{c}{CNN(L)}&\multicolumn{1}{c}{CapsNet(L)}
        &\multicolumn{1}{c}{(L)}&\multicolumn{1}{c}{(L)}\\
        \hline
        1 & 20.23& 7.41 & 39.61&43.87& 58.97&\multicolumn{1}{c|}{\textbf{52.30}} & \multicolumn{1}{c}{37.23}& \multicolumn{1}{c}{11.38}& \multicolumn{1}{c}{87.30} &90.31& \textbf{91.41}&88.84\\
        2 & 15.51& 2.88 & 22.03&26.69& 26.00&\multicolumn{1}{c|}{\textbf{35.40}}& \multicolumn{1}{c}{61.73}& \multicolumn{1}{c}{68.27}& \multicolumn{1}{c}{88.93} &91.35& \textbf{95.64}&89.34\\
        3 & 40.20& 26.40& 68.37&\textbf{82.57}& 28.98&\multicolumn{1}{c|}{46.60}& \multicolumn{1}{c}{50.73}& \multicolumn{1}{c}{20.04}& \multicolumn{1}{c}{62.00} &72.65& 57.41&\textbf{73.76}\\
        4 & \textbf{94.98}& 37.69& 81.81&66.48& 74.59&\multicolumn{1}{c|}{79.83}& \multicolumn{1}{c}{69.18}& \multicolumn{1}{c}{66.58}& \multicolumn{1}{c}{\textbf{99.28}} &93.23& 97.66&94.25\\
        5 & 26.42& 13.79& \textbf{60.74}&38.83& 24.87&\multicolumn{1}{c|}{44.92}& \multicolumn{1}{c}{28.80}& \multicolumn{1}{c}{47.91}& \multicolumn{1}{c}{65.32} &83.07& 68.61&\textbf{86.93}\\
        6 & \textbf{60.14}& 55.94& 21.40&31.47& 37.53&\multicolumn{1}{c|}{52.91}& \multicolumn{1}{c}{71.80}& \multicolumn{1}{c}{24.20}& \multicolumn{1}{c}{81.72} &61.85& 84.95&\textbf{85.90}\\
        7 & 41.60  & 38.62& \textbf{80.15}&55.50& 88.22&\multicolumn{1}{c|}{79.20}\\
        8 & 65.81  & 75.21& 65.96         &86.80& 75.34&\multicolumn{1}{c|}{\textbf{92.47}}\\
        9 & 11.80  & 11.49& \textbf{60.90}&40.42& 54.04&\multicolumn{1}{c|}{51.15}\\
        10 & 9.74  & 7.01 & 41.54         &50.29& \textbf{58.72}&\multicolumn{1}{c|}{46.07}\\
        11 & 35.58 & 35.96& 74.20         &66.70& 82.64&\multicolumn{1}{c|}{\textbf{89.25}}\\
        12 & 12.10 & 3.33 & 41.62&\textbf{42.75}& 13.74&\multicolumn{1}{c|}{38.52}\\
        13 & 31.23 & 47.49& 49.70         &51.23& 74.50&\multicolumn{1}{c|}{\textbf{61.29}}\\
        14 & 76.92 & 27.6 & 39.24&\textbf{85.83}& 53.85&\multicolumn{1}{c|}{54.12}\\
        15 & 3.81  & 14.59& 39.81         &\textbf{67.86}& 20.79&\multicolumn{1}{c|}{53.63}\\
        \hline
        OA & 33.71         & 24.21& 53.61&54.15& 53.50&\multicolumn{1}{c|}{\textbf{59.83}}& \multicolumn{1}{c}{50.15}& \multicolumn{1}{c}{47.69}& \multicolumn{1}{c}{82.45}&85.50& 84.56&\textbf{89.31}\\
        AA & 36.41         & 27.03& 52.47&55.82& 51.52&\multicolumn{1}{c|}{\textbf{58.49}}& \multicolumn{1}{c}{53.24}& \multicolumn{1}{c}{39.73}& \multicolumn{1}{c}{80.76}&82.08& 82.62&\textbf{86.51}\\
        \(\kappa\) & 28.63 & 19.26& 49.80&50.55& 49.68&\multicolumn{1}{c|}{\textbf{56.48}}& \multicolumn{1}{c}{38.48}& \multicolumn{1}{c}{32.28}& \multicolumn{1}{c}{77.24}&80.99& 79.98&\textbf{85.98}\\
        \specialrule{0.1em}{0.5pt}{0.5pt}
    \end{tabular}
\end{table*}

On the basis of the parameter settings reported in subsection IV-B, the experimental results obtained by all the considered methods on the HSI data alone, the LiDAR data alone, and HSI + LiDAR data are reported in Tables VII-X. For convenience, HSI data, LiDAR data, and HSI + LiDAR data in Tables VII-X are abbreviated as H, L, and H+L, respectively. Particularly, for SSCL3DNN (which merges H+L) in Tables VIII-IX, the input of SSCL3DNN \cite{WH2019} is changed to the fusion of the HSI and LiDAR data, and the SaCL2DNN model \cite{WH2019} is selected for extracting spatial features from the LiDAR data, since SSCL3DNN is not suitable for dealing with 2-D data (and also because SaCL2DNN has a similar structure with regards to SSCL3DNN). From Tables VII-X, it can be seen that the proposed dual-channel $A^{3}$CLNN can obtain better classification performance than the other tested methods. On the one hand, the gate mechanisms realized by the convolution operation make it possible for the ConvLSTM2D and ConvLSTM3D layers to fully exploit both spatial and spectral information from multisource remote sensing data. On the other hand, with the help of the spectral, spatial and multiscale attention mechanisms, the proposed model can extract highly effective spectral- and spatial-enhanced features, and fully exploit multiscale information coming from multisource remote sensing data. Furthermore, the three-level fusion and stepwise training strategies not only can fully integrate the spectral and spatial information by exploiting the complementary information of HSI and LiDAR data, but also accelerate the convergence speed of the whole model. Concretely, the experimental results obtained when analyzing the effectiveness of the proposed three-level fusion strategy and stepwise training strategy are given in Table VII, from which it is apparent that, for the LiDAR data in Houston and Trento data sets, the values of the OA metric in the LiDAR branch are 59.83\% and 89.31\%, respectively, while the values of the OA metric in the HSI branch are 87.00\% and 97.65\%, respectively. Moreover, when utilizing HSI and LiDAR data, the gains in OA obtained by using our newly developed model are respectively 3.55\% and 1.08\% (compared to the HSI channel) for the two considered data sets. This demonstrates the effectiveness of the proposed dual-channel $A^{3}$CLNN.

Moveover, the experimental results aimed at analyzing the accuracy obtained for each class and quantitative metrics obtained after fusing HSI and LiDAR data are reported in Tables VIII-X. From these tables it can be seen that, without data augmentation, the proposed dual-channel $A^{3}$CLNN model performs better than other baseline methods. Concretely, for the Houston data set, the proposed model yields highly competitive classification accuracy of 90.55\%, with a gain over 7.40\% with respect to that achieved by the two-branch CNN. Our model also yields 10.40\% and 9.79\% improvements with regards to the standard SVM and ELM, respectively. As for the Trento data set, the improvements in OA achieved by the proposed dual-channel $A^{3}$CLNN model are 6.04\%, 7.88\%, and 3.37\%, respectively, when compared with SVM, ELM, and two-branch CNN. For dual-channel CapsNet \cite{HL2020}, although many works have verified that CapsNet can better learn the information of position, orientation, deformation, and texture than CNN, the spectral and scale information contained in different classes may not be fully learned. The proposed model outperforms dual-channel CapsNet, having improvements of 3.94\% and 1.98\% for the two data sets, respectively. The above experimental results show that, for traditional models such as SVM and ELM, the way of converting the HSI and LiDAR data into vectors leads to the loss of the spatial and geometric structure information, while simple feature cascading in two-branch CNN \cite{XX2018} and dual-channel CapsNet \cite{HL2020} fails to effectively learn complementary information by fusing the HSI and LiDAR data. In contrast to these comparison algorithms, the design of gate mechanisms makes the ConvLSTM-based models to fully leverage spatial information and better preserve the intrinsic structure information of the original data, which is in line with the findings in \cite{WH2019}. In addition, compared with SSCL3DNN, our dual-channel $A^{3}$CLNN model can improve the classification accuracy in almost all cases, and obtain 4.54\% and 2.27\% gains in OA for the two data sets, respectively, benefiting from the developed three-level fusion and stepwise training strategies. In addition, when only HSI or LiDAR data are used, the performance of each branch in the proposed model is also superior to that achieved by other comparison methods. Specifically, the dual-channel $A^{3}$CLNN(H) model can even achieve higher accuracy than other models that use both HSI and LiDAR data, which may be the contributions of the spectral and multiscale attention structures. More detailed experimental results can be found in Tables VIII-X.

Regarding Tables VIII-IX, similar conclusions can be drawn from the classification maps presented in Figs. 8-9, from which it is obvious that the classification maps yielded by the proposed dual-channel $A^{3}$CLNN model are the closest to the ground-truth maps for the two considered data sets. Particularly, the obtained classification maps exhibit less mislabeled areas, and the boundaries between different classes are better delineated and identified, especially for classes 6, 8 and 11 in Fig. 8, and classes 2, 3 and 5 in Fig. 9. This further verifies the advantages and effectiveness of our dual-channel $A^{3}$CLNN model.

\subsection*{D. Ablation Study}
To highlight the effectiveness of SeAB, SaAB, MSRAB, two-level attention strategy, and stepwise training strategy in dual-channel $A^{3}$CLNN, detailed ablation studies are conducted to see how they contribute to the classification performance. SSCL3DNN \cite{WH2019} is used as a baseline. For convenience, the dual-channel $A^{3}$CLNN model with and without each component are respectively abbreviated as proposed($\cdot$, with) and proposed($\cdot$, without), in which $\cdot$ denotes H, L, and H+L.
\subsubsection*{(1) The effectiveness of SeAB} To effectively learn the spectral-enhanced feature representation from the HSI data, a SeAB module is built for the classification of the HSI branch. To demonstrate its effectiveness, we conduct experiments to analyze the influence of SeAB by adding and removing it from our dual-channel $A^{3}$CLNN, whose results are reported in Table XI. Compared with SSCL3DNN(H), the proposed model(H, without SeAB) can obtain 3.13\% and 1.57\% gains for the two data sets, respectively, while the proposed model(H+L, without SeAB) generates 4.04\% and 1.65\% improvements against SSCL3DNN(H+L). With the help of SeAB, the proposed model(H+L) further improves the classification accuracy of the proposed model(H+L, without SeAB) by 0.50\% and 0.62\% for the two data sets, indicating that the SeAB module can improve the classification performance by enhancing the ability of spectral feature representation of the whole model.
\subsubsection*{(2) The structure analysis of SaAB} LiDAR data can provide rich elevation information in the spatial domain, thus having the potential of improving the characterization of HSI scenes. The experimental results for analyzing the performance of our dual-channel $A^{3}$CLNN model with and without SaAB are reported in Table XII. Compared with the proposed model(H+L, without SaAB), 1.22\% and 0.56\% gains are yielded by the proposed model(H+L, with SaAB) for the two data sets, respectively, which shows the advantages of SaAB.
\subsubsection*{(3) The analysis of two-level attention strategy} To fully utilize and fuse the spectral and spatial information, a composite attention learning module is designed as a two-level attention strategy to jointly learn spectral- and spatial-enhanced features. The experiments for analyzing its influence on the classification performance are carried out, whose results are presented in Table XIII. Compared with SSCL3DNN(H+L), the gains in OA yielded by the proposed model(H+L, without composite attention learning) are respectively 2.99\% and 1.17\% for the two data sets, showing the superiority of the MSRAB model to improve the classification performance to some extent. Furthermore, after embedding the composite attention learning module, the proposed model(H+L) achieves 1.55\% and 1.10\% gains for the two data sets, respectively. Experimental results in Table XIII show that the composite attention learning and MSRAB modules can effectively learn the spatial-spectral and multiscale information, resulting in better classification results.
\begin{table}[H]
    \centering
    \renewcommand\thetable{\Roman{table}}
    \renewcommand\tabcolsep{3.0pt}
    \caption{The Influence of The Proposed SeAB Module}
    \scriptsize
    \begin{tabular}{p{4.06cm}<{\centering}p{2cm}<{\centering}}
    \specialrule{0.1em}{0.5pt}{0.5pt}
      $ \begin{tabular}{c|cc|}
        \multicolumn{1}{c}{\multirow{2}*{Models}} \quad& \multicolumn{2}{|c|}{Houston Data}\\
        \quad& \multicolumn{1}{c}{OA} \quad& \multicolumn{1}{c|}{Kappa}\\
        \hline
        SSCL3DNN(H)           & 82.72         &81.33\\
        Proposed(H, without)  & 85.85         &84.64\\
        Proposed(H, with)     & \textbf{87.00}&\textbf{85.90}\\
        \hline
        SSCL3DNN(H+L)         & 86.01         &84.84\\
        Proposed(H+L, without)& 90.05         &89.22\\
        Proposed(H+L, with)   & \textbf{90.55}&\textbf{89.75}\\
      \end{tabular}$ &
      $ \begin{tabular}{c c c}
        \multicolumn{3}{c}{Trento Data}\\
        \quad& OA \quad& Kappa\\
        \hline
        & 95.50          & 93.99          \\
        & 97.07          & 96.10          \\
        & \textbf{97.65} & \textbf{96.86} \\
        \hline
        & 96.46          & 95.30          \\
        & 98.11          & 97.47          \\
        & \textbf{98.73} & \textbf{98.31} \\
      \end{tabular}$ \\
    \specialrule{0.1em}{0.5pt}{0.5pt}
    \end{tabular}
\vspace{-0.6cm}
\end{table}
\begin{table}[H]
    \centering
    \renewcommand\thetable{\Roman{table}}
    \renewcommand\tabcolsep{3.0pt}
    \caption{The Structure Analysis of The Developed SaAB Module}
    \scriptsize
    \begin{tabular}{p{4.06cm}<{\centering}p{2cm}<{\centering}}
    \specialrule{0.1em}{0.5pt}{0.5pt}
      $ \begin{tabular}{c|cc|}
        \multicolumn{1}{c}{\multirow{2}*{Models}} \quad& \multicolumn{2}{|c|}{Houston Data}\\
        \quad& \multicolumn{1}{c}{OA} \quad& \multicolumn{1}{c|}{Kappa}\\
        \hline
        SaCL2DNN(L)            & 53.50         &49.68\\
        Proposed(L, without)   & 58.48         &55.05\\
        Proposed(L, with)      & \textbf{59.83}&\textbf{56.48}\\
        \hline
        SSCL3DNN(H+L)          & 86.01         &84.84\\
        Proposed(H+L, without) & 89.33         &88.45\\
        Proposed(H+L, with)    & \textbf{90.55}&\textbf{89.75}\\
      \end{tabular}$ &
      $ \begin{tabular}{c c c}
        \multicolumn{3}{c}{Trento Data}\\
        \quad& OA \quad& Kappa\\
        \hline
        & 84.56          & 79.98          \\
        & 87.59          & 83.60          \\
        & \textbf{89.31} & \textbf{85.98} \\
        \hline
        & 96.46          & 95.30          \\
        & 98.17          & 97.56          \\
        & \textbf{98.73} & \textbf{98.31} \\
      \end{tabular}$ \\
    \specialrule{0.1em}{0.5pt}{0.5pt}
    \end{tabular}
\vspace{-0.6cm}
\end{table}
\begin{table}[H]
    \centering
    \renewcommand\thetable{\Roman{table}}
    \renewcommand\tabcolsep{3.0pt}
    \caption{The Analysis of Two-Level Attention Strategy}
    \scriptsize
    \begin{tabular}{p{4.06cm}<{\centering}p{2cm}<{\centering}}
    \specialrule{0.1em}{0.5pt}{0.5pt}
      $ \begin{tabular}{c|cc|}
        \multicolumn{1}{c}{\multirow{2}*{Models}} \quad& \multicolumn{2}{|c|}{Houston Data}\\
        \quad& \multicolumn{1}{c}{OA} \quad& \multicolumn{1}{c|}{Kappa}\\
        \hline
        SSCL3DNN(H+L)          & 86.01         &84.84\\
        Proposed(H+L, without) & 89.00         &88.11\\
        Proposed(H+L, with)    & \textbf{90.55}&\textbf{89.75}\\
      \end{tabular}$ &
      $ \begin{tabular}{c c c}
        \multicolumn{3}{c}{Trento Data}\\
        \quad& OA \quad& Kappa\\
        \hline
        & 96.46          & 95.30          \\
        & 97.63          & 96.83          \\
        & \textbf{98.73} & \textbf{98.31} \\
      \end{tabular}$ \\
    \specialrule{0.1em}{0.5pt}{0.5pt}
    \end{tabular}
\vspace{-0.8cm}
\end{table}
\begin{table}[H]
    \centering
    \renewcommand\thetable{\Roman{table}}
    \renewcommand\tabcolsep{3.0pt}
    \caption{The Architecture Study of The Designed MSRAB Module}
    \scriptsize
    \begin{tabular}{p{4.06cm}<{\centering}p{2cm}<{\centering}}
    \specialrule{0.1em}{0.5pt}{0.5pt}
      $ \begin{tabular}{c|cc|}
        \multicolumn{1}{c}{\multirow{2}*{Models}} \quad& \multicolumn{2}{|c|}{Houston Data}\\
        \quad& \multicolumn{1}{c}{OA} \quad& \multicolumn{1}{c|}{Kappa}\\
        \hline
        SSCL3DNN(H)             & 82.72         &81.33\\
        Proposed(H, without)    & 85.29         &84.08\\
        Proposed(H, with)       & \textbf{87.00}&\textbf{85.90}\\
        \hline
        SaCL2DNN(L)             & 53.50         &49.68\\
        Proposed(L, without)    & 56.34         &52.73\\
        Proposed(L, with)       & \textbf{59.83}&\textbf{56.48}\\
        \hline
        SSCL3DNN(H+L)           & 86.01         &84.84\\
        Proposed(H+L, without)  & 87.23         &86.23\\
        Proposed(H+L, with)     & \textbf{90.55}&\textbf{89.75}\\
      \end{tabular}$ &
      $ \begin{tabular}{c c c}
        \multicolumn{3}{c}{Trento Data}\\
        \quad& OA \quad& Kappa\\
        \hline
        & 95.50          & 93.99          \\
        & 96.33          & 95.11          \\
        & \textbf{97.65} & \textbf{96.86} \\
        \hline
        & 84.56          & 79.98          \\
        & 87.81          & 84.16          \\
        & \textbf{89.31} & \textbf{85.98} \\
        \hline
        & 96.46          & 95.30          \\
        & 97.58          & 96.77          \\
        & \textbf{98.73} & \textbf{98.31} \\
      \end{tabular}$ \\
    \specialrule{0.1em}{0.5pt}{0.5pt}
    \end{tabular}
\vspace{-0.4cm}
\end{table}

\subsubsection*{(4) The architecture study of MSRAB} Different classes in hyperspectral data should contain different scale information, so that using the fixed-scale convolution kernel limits the ability of CNN-based models to learn scale information. To analyze the contributions of MSRAB, we compare the proposed model with a variant of our model in which the MSRAB module is replaced with a ConvLSTM2D or ConvLSTM3D layer in Table XIV. We can observe that, compared with this variant, MSRAB can bring 3.32\% and 1.15\% improvements to the proposed model(H+L) for the two data sets, respectively. Moreover, the experimental results shown in the last three lines of Table XIV also validate that the composite attention learning and MSRAB modules are one of the main reasons for the performance improvements of dual-channel $A^{3}$CLNN, which is consistent with the conclusions in Table XIII. More detailed experimental results are reported in Table XIV.
\begin{figure}[H]
    \centering
    \renewcommand\tabcolsep{1.0pt}
    \vspace{-0.2cm}
    \scriptsize 
    \begin{tabular}{p{4.1cm}<{\centering}p{4cm}<{\centering}}
        \begin{minipage}[t]{1\linewidth}
            \centering
            \includegraphics[width=1.55in]{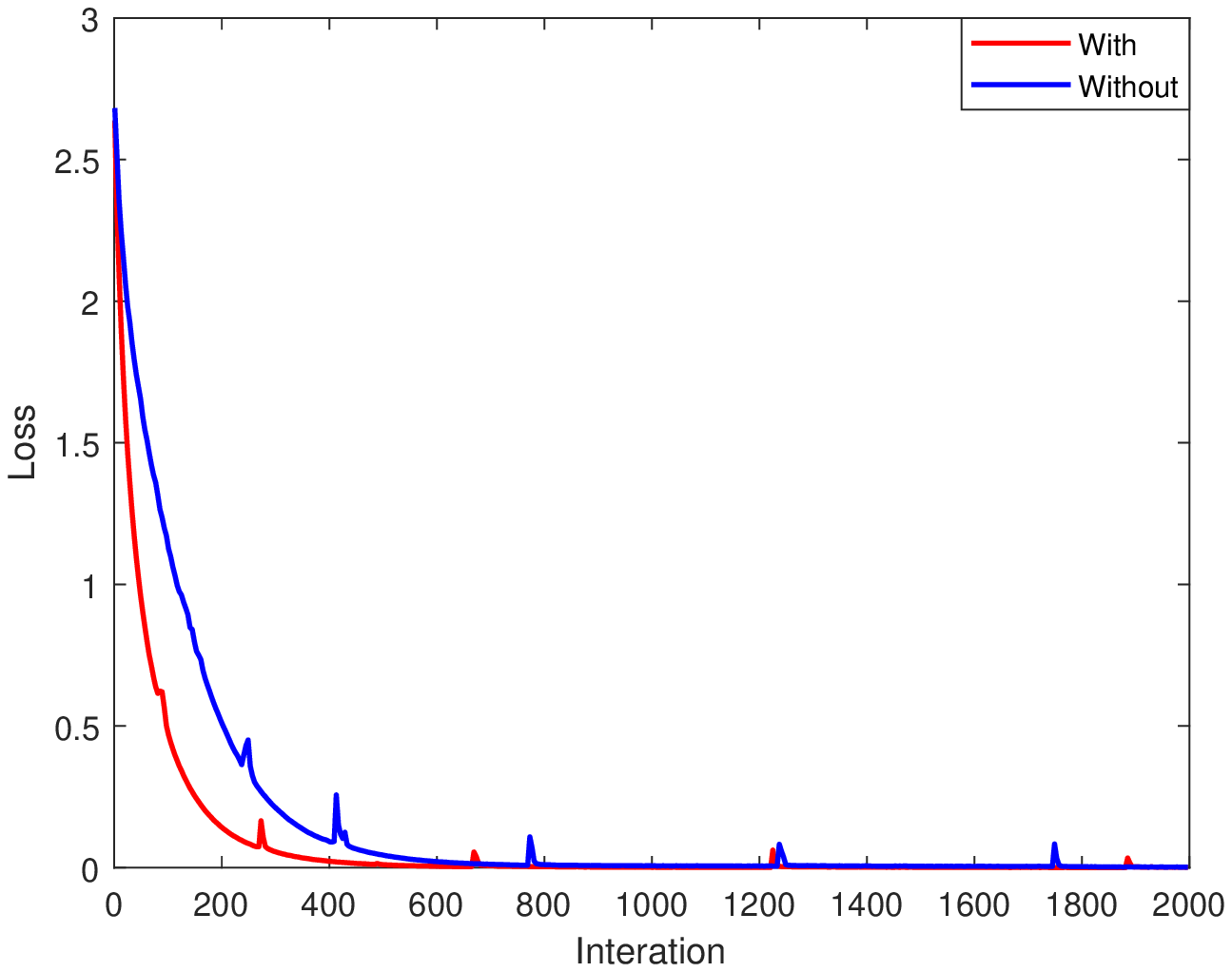}
        \end{minipage}
        & \begin{minipage}[t]{1\linewidth}
            \centering
            \includegraphics[width=1.55in]{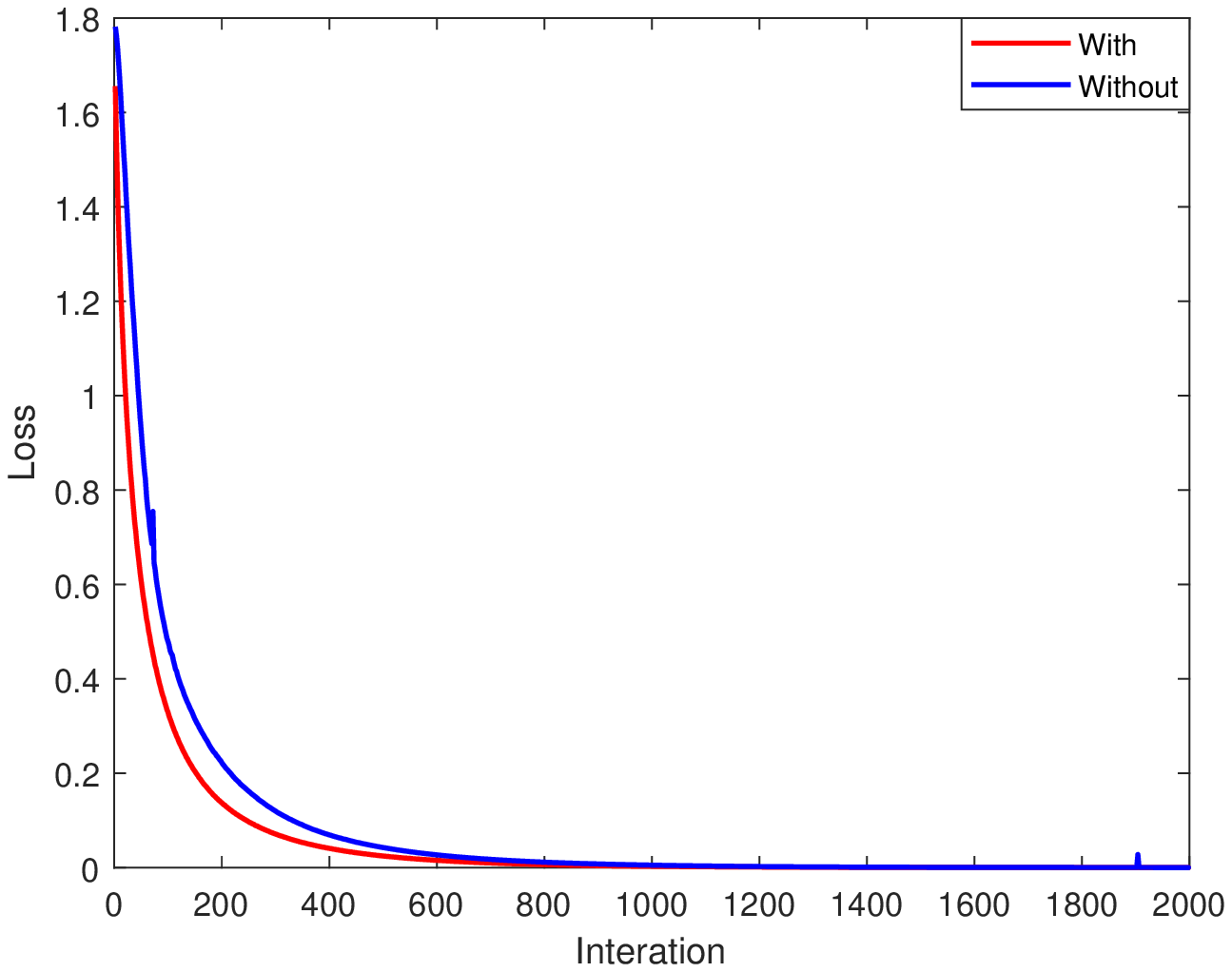}
        \end{minipage}\\
        (a) & (b) \\
    \end{tabular}
\caption{The learning curves of the proposed dual-channel $A^{3}$CLNN for the Houston and Trento data sets. (a) The learning curves for the Houston data set. (b) The learning curves for the Trento data set.}
\vspace{-0.2cm}
\end{figure}

\subsubsection*{(5) The analysis of stepwise training strategy} To obtain better optimization of our dual-channel $A^{3}$CLNN, a stepwise training strategy is designed for the full training of the whole model. The learning curves of the proposed method on the two data sets are visualized in Fig. 10, where both curves for each data set tend to converge stably with the increase of the number of iterations. By analyzing the data distribution of the two data sets in Figs. 5-6, we can find that the data distribution of the Houston data set is more scattered than that of the Trento data set. From Fig. 10(a), although these two curves fluctuate slightly, the loss curve of the proposed model with stepwise training strategy converges faster and is more stable than the one without it. The learning curves shown in Fig. 10 illustrate that the proposed stepwise training strategy is advantageous for accelerating the convergence speed of the proposed dual-channel $A^{3}$CLNN model to some extent.

\section{Conclusion}
In this paper, a new dual-channel $A^{3}$CLNN model has been proposed for the classification of multisource remote sensing data. Specifically, our model comprises two different pipelines for LiDAR data and HSI data, in which spatial, spectral, and multiscale residual attention structures have been implemented to fully exploit spatial, spectral, and multiscale information and obtain more comprehensive and discriminative feature representation. Moreover, an effective three-level fusion strategy and a novel stepwise training strategy are also developed to fully integrate the spatial and spectral information contained in the LiDAR and HSI data, exploiting their complementarity. Our experimental results demonstrate that the proposed dual-channel $A^{3}$CLNN provides better performance than state-of-the-art CNN-based approaches (e.g., the two-branch CNN), the capsule network-based models (e.g., the dual-channel CapsNet) and baseline ConvLSTM-based methods (e.g., SSCL3DNN).

%
%

\ifCLASSOPTIONcaptionsoff
  \newpage
\fi

\end{spacing}
\end{document}